\documentclass{article}

\usepackage[numbers,compress,sort]{natbib}

\usepackage{neurips_data_2024}

\usepackage{booktabs}
\usepackage{multirow}
\usepackage{graphicx}
\usepackage[dvipsnames]{xcolor}
\usepackage{siunitx}
\usepackage{rotating}
\usepackage{caption}
\usepackage{changepage}
\usepackage{placeins}
\usepackage{wrapfig}
\usepackage{bm}
\usepackage{bibunits}

\usepackage{colortbl}
\definecolor{1}{rgb}{0,0.5,0} 
\definecolor{2}{rgb}{0,0.75,0} 
\definecolor{3}{rgb}{0.5,1,0.5} 

\usepackage[utf8]{inputenc} 
\usepackage[T1]{fontenc}    
\usepackage{hyperref}       
\usepackage{url}            
\usepackage{booktabs}       
\usepackage{amsfonts}       
\usepackage{nicefrac}       
\usepackage{microtype}      
\usepackage{xcolor}         
\usepackage{amsmath}
\usepackage{caption}
\usepackage{wrapfig}
\usepackage{makecell}
\usepackage{subcaption}
\usepackage{tikz}
\usepackage{fontawesome}
\usepackage[sectionbib]{chapterbib}

\newcommand{\hy}[1]{\textcolor{OliveGreen}{#1}}
\newcommand{\pan}[1]{\textcolor{blue}{P:#1}}

\newcommand*\circled[1]{\raisebox{.4pt}
                    {\tikz[baseline=(char.base)]{
            \node[shape=circle,draw,inner sep=1pt, style={fill=white, text=black}, scale=0.75] (char) {\textbf{#1}};}}}

\title{A Benchmark on Directed Graph Representation Learning in Hardware Designs}

\author{%
  Haoyu Wang \\
  Georgia Tech
  \And
 Yinan Huang \\
 Georgia Tech
  \And
  Nan Wu \\
  George Washington University
  \And
  Pan Li \\
  Georgia Tech
}

\begin{document}
\newcommand{\defeq}{\stackrel{\tiny\mathrm{def}}{=}}
\newcommand{\dd}{\mathrm{d}}
\newcommand{\thalf}{\tfrac{1}{2}}

\newcommand{\iid}{\stackrel{\tiny\mathrm{iid}}{\sim}}
\newcommand{\E}{\operatorname{\mathsf E}}
\newcommand{\Var}{\operatorname{\mathsf Var}}
\renewcommand{\H}{\operatorname{\mathsf H}}
\newcommand{\I}{\operatorname{\mathsf I}}
\newcommand{\D}{\operatorname{\mathsf D}}
\newcommand{\Cov}{\operatorname{\mathsf Cov}}
\newcommand{\FIM}{\operatorname{\mathcal{F}}}

\newcommand{\R}{\operatorname{\mathbb R}}
\newcommand{\Z}{\operatorname{\mathbb Z}}
\newcommand{\X}{\operatorname{\mathcal X}}
\newcommand{\Y}{\operatorname{\mathcal Y}}

\newcommand{\MVN}{\operatorname{MVN}}
\newcommand{\Normal}{\operatorname{Normal}}
\newcommand{\Beta}{\operatorname{Beta}}
\newcommand{\Bernoulli}{\operatorname{Bernoulli}}
\newcommand{\Categorical}{\operatorname{Categorical}}
\newcommand{\Shuffle}{\operatorname{Shuffle}}

\newcommand{\sign}{\operatorname{sign}}
\newcommand{\logit}{\operatorname{logit}}
\newcommand{\expit}{\operatorname{expit}}
\newcommand{\softmax}{\operatorname{softmax}}
\newcommand{\softplus}{\operatorname{softplus}}
\newcommand{\logsumexp}{\operatorname{logsumexp}}

\newcommand{\tee}{\mathsf{\tiny T}}
\newcommand{\diag}{\operatorname{diag}}
\newcommand{\trace}{\operatorname{trace}}
\newcommand{\chol}{\operatorname{chol}}
\newcommand{\zeros}{\operatorname{\mathbb{0}}}
\newcommand{\ones}{\operatorname{\mathbb{1}}}
\newcommand{\eye}{\operatorname{I}}

\newcommand{\Encoder}{\operatorname{Encoder}}
\newcommand{\Decoder}{\operatorname{Decoder}}
\newcommand{\MLP}{\operatorname{MLP}}
\newcommand{\ResNet}{\operatorname{ResNet}}
\newcommand{\Augment}{\operatorname{Augment}}

\renewcommand{\dot}[2]{\left\langle #1, #2 \right\rangle}
\newcommand{\norm}[1]{\left\lVert #1 \right\rVert}
\newcommand{\abs}[1]{\left| #1 \right|}
\newcommand{\ceil}[1]{\left\lceil #1 \rceil|}
\newcommand{\floor}[1]{\left\lfloor #1 \rfloor|}

\newcommand{\domain}{\operatorname{domain}}
\newcommand{\range}{\operatorname{range}}
\newcommand{\ndims}{\operatorname{ndims}}
\newcommand{\shape}{\operatorname{shape}}
\newcommand{\size}{\operatorname{size}}
\newcommand{\len}{\operatorname{len}}
\newcommand{\indexes}{\operatorname{indexes}}

\newtheorem{thm}{Theorem}[section]
\newtheorem{lma}[thm]{Lemma}
\newtheorem{prop}[thm]{Proposition}
\newtheorem{cor}[thm]{Corollary}
\newtheorem{clm}[thm]{Claim}

\newtheorem{defn}[thm]{Definition}
\newtheorem{examp}[thm]{Example}
\newtheorem{conj}[thm]{Conjecture}
\newtheorem{rmk}[thm]{Remark}

\def\vzero{{\bm{0}}}
\def\vone{{\bm{1}}}
\def\vmu{{\bm{\mu}}}
\def\vtheta{{\bm{\theta}}}
\def\va{{\bm{a}}}
\def\vb{{\bm{b}}}
\def\vc{{\bm{c}}}
\def\vd{{\bm{d}}}
\def\ve{{\bm{e}}}
\def\vf{{\bm{f}}}
\def\vg{{\bm{g}}}
\def\vh{{\bm{h}}}
\def\vi{{\bm{i}}}
\def\vj{{\bm{j}}}
\def\vk{{\bm{k}}}
\def\vl{{\bm{l}}}
\def\vm{{\bm{m}}}
\def\vn{{\bm{n}}}
\def\vo{{\bm{o}}}
\def\vp{{\bm{p}}}
\def\vq{{\bm{q}}}
\def\vr{{\bm{r}}}
\def\vs{{\bm{s}}}
\def\vt{{\bm{t}}}
\def\vu{{\bm{u}}}
\def\vv{{\bm{v}}}
\def\vw{{\bm{w}}}
\def\vx{{\bm{x}}}
\def\vy{{\bm{y}}}
\def\vz{{\bm{z}}}
\def\tV{{\tens{V}}}
\def\tW{{\tens{W}}}
\def\tX{{\tens{X}}}
\def\tY{{\tens{Y}}}
\def\tZ{{\tens{Z}}}

\maketitle
\renewcommand{\thefootnote}{}
\footnotetext{ Emails: haoyu.wang@gatech.edu, \ panli@gatech.edu}
\renewcommand{\thefootnote}{\arabic{footnote}}

\begin{abstract}
To keep pace with the rapid advancements in design complexity within modern computing systems, directed graph representation learning (DGRL) has become crucial, particularly for encoding circuit netlists, computational graphs, and developing surrogate models for hardware performance prediction. However, DGRL remains relatively unexplored, especially in the hardware domain, mainly due to the lack of comprehensive and user-friendly benchmarks. This study presents a novel benchmark comprising five hardware design datasets and 13 prediction tasks spanning various levels of circuit abstraction. We evaluate 21 DGRL models, employing diverse graph neural networks and graph transformers (GTs) as backbones, enhanced by positional encodings (PEs) tailored for directed graphs. Our results highlight that bidirected (BI) message passing neural networks (MPNNs) and robust PEs significantly enhance model performance. Notably, the top-performing models include PE-enhanced GTs interleaved with BI-MPNN layers and BI-Graph Isomorphism Network, both surpassing baselines across the 13 tasks. 
Additionally, our investigation into out-of-distribution (OOD) performance emphasizes the urgent need to improve OOD generalization in DGRL models. This benchmark, implemented with a modular codebase, streamlines the evaluation of DGRL models for both hardware and ML practitioners.\footnote{Document for the toolbox is available at: \url{https://benchmark-for-dgrl-in-hardwares.readthedocs.io/en/latest/}.}

\end{abstract}

\section{Introduction}
Directed graphs, where edges encode directional information, are widely utilized as data models in various applications, including email communication~\cite{kossinets2008structure, khrabrov2010discovering}, financial transactions~\cite{gale2007financial, chinazzi2015systemic,tiwari2021network}, and supply chains~\cite{surana2005supply, kaur2006graph, wagner2010assessing}. 
Notably, hardware designs can be represented as directed graphs, such as circuit netlists~\cite{hachtel2005logic,vladimirescu1994spice}, control and data flow graphs~\cite{cummins2021a,wu2022high,bai2023towards, ye2024hida}, or computational graphs~\cite{zhang2021nn,phothilimthana2024tpugraphs}, often exhibiting unique properties.
These graph structures reflect restricted connection patterns among circuit components or program operation units, with directed edges encapsulating long-range directional and logical dependencies.

Recently, employing machine learning (ML) to assess the properties of hardware designs via their directed graph representations has attracted significant attention~\cite{wu2022survey,bai2023towards,dong2023cktgnn,li2020customized,ma2019high,bucher2022appgnn,he2021graph,guo2022timing,phothilimthana2024tpugraphs}.
Traditional simulation-based methods often require considerable time (hours or days) to achieve the desired accuracy in assessing design quality~\cite{zhao2017comba,dai2018fast,wu2021ironman,wu2022high}, substantially slowing down the hardware development cycle due to repeated optimization-evaluation iterations. 
In contrast, ML models can serve as faster and more cost-effective surrogates for simulators, offering a balanced alternative between simulation costs and prediction accuracy~\cite{mirhoseini2021graph,al2021deep,wu2021ironman,chen2018learning, jia2020improving, cakir2018reverse,dudziak2020brp, cao2022domain, liu2021parasitic, wang2020gcn, wu2023gamora}.
Such an approach is promising to expedite hardware evaluation, especially given the rapid growth of design complexity in modern electronics and computing systems~\cite{roadmap}.

Despite the promising use cases, developing ML models for reliable predictions on directed graphs, particularly within hardware design loops, is still in its early stages, largely due to the lack of comprehensive and user-friendly benchmarks. Existing studies in the ML community have primarily focused on undirected graphs, utilizing Graph Neural Networks (GNNs)~\cite{kipf2016semi, xu2018powerful, velivckovic2017graph} or Graph Transformers (GTs)~\cite{rampavsek2022recipe, kreuzer2021rethinking, ying2021transformers,min2022transformer}. Among the limited studies on directed graph representation learning (DGRL)~\cite{zhang2021magnet, tong2020directed, tong2020digraph, geisler2023transformers}, most have only evaluated their models for node/link-level predictions on single graphs in domains such as web networks, or financial networks~\cite{he2024pytorch}. These domains exhibit very different connection patterns compared to those in hardware design. To the best of our knowledge, CODE2 in the Open Graph Benchmark (OGB)~\cite{hu2020open} is the only commonly used benchmark that may share some similarities with hardware data. However, the graphs in CODE2 are IRs of Python programs, which may not fully reflect the properties of data in hardware design loops.

Numerous DGRL models for hardware design tasks have been developed by domain experts. While promising, hardware experts tend to incorporate domain-specific insights with off-the-shelf GNNs (e.g., developing hierarchical GNNs to mimic circuit modules~\cite{wu2022high,dong2023cktgnn} or encoding circuit fan-in and fan-out in node features~\cite{ren2020paragraph,gnn_re,vasudevan2021learning}), with limited common design principles investigated in model development.
In contrast, state-of-the-art (SOTA) DGRL techniques proposed by the ML community lack thorough investigation in these tasks. These techniques potentially offer a more general and effective manner of capturing data patterns that might be overlooked by domain experts.


\textbf{Present Benchmark.} This work addresses the aforementioned gaps by establishing a new benchmark consisting of representative hardware design tasks and extensively evaluating various DGRL techniques for these tasks. On one hand, the evaluation results facilitate the identification of commonly useful principles for DGRL in hardware design. On the other hand, the ML community can leverage this benchmark to further advance DGRL techniques.

\begin{figure}[t]
    \centering
    \includegraphics[width = \textwidth]{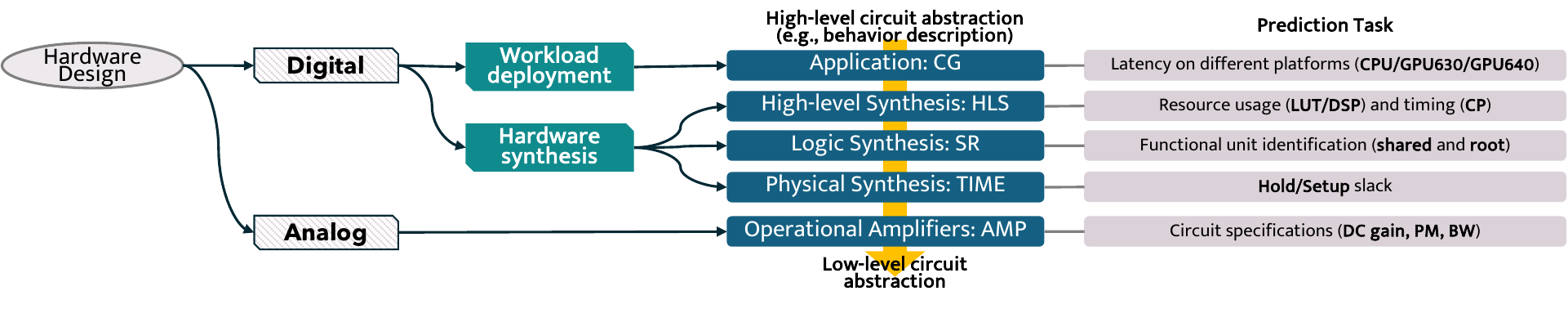}
    \vspace{-0.7cm}
    \caption{Coverage of Datasets/Tasks.}
    \label{fig:representativeness_datasets}
\end{figure}

Specifically, our benchmark collects five hardware design datasets encompassing a total of 13 prediction tasks. The data spans different levels of circuit abstraction, with graph sizes reaching up to 400+ nodes per graph across 10k+ graphs for graph-level tasks, and up to 50k+ nodes per graph for node-level tasks (see Fig.~\ref{fig:representativeness_datasets} and Table.~\ref{tab:datasets}). We also evaluate 21 DGRL models based on 8 GNN/GT backbones, combined with different message passing directions and various enhancements using positional encodings (PEs) for directed graphs~\cite{geisler2023transformers}. PEs are vectorized representations of node positions in graphs and have been shown to improve the expressive power of GT/GNNs for undirected graphs~\cite{wang2022equivariant,huang2023stability,lim2022sign, rampavsek2022recipe}. PEs for directed graphs are still under-explored~\cite{geisler2023transformers}, but we believe they could be beneficial for hardware design tasks that involve long-range and logical dependencies.

Our extensive evaluations provide significant insights into DGRL for hardware design tasks. Firstly, bidirected (BI) message passing neural networks (MPNNs) can substantially improve performance for both pure GNN encoders and GT encoders that incorporate MPNN layers, such as GPS~\cite{rampavsek2022recipe}. Secondly, PEs, only when used stably~\cite{wang2022equivariant, huang2023stability}, can broadly enhance the performance of both GTs and GNNs. This observation contrasts with findings from undirected graph studies, particularly in molecule property prediction tasks, where even unstable uses of PEs may improve model performance~\cite{dwivedi2020benchmarking, kreuzer2021rethinking,lim2022sign, rampavsek2022recipe}. Thirdly, GTs with MPNN layers typically outperform pure GNNs on small graphs but encounter scalability issues when applied to larger graphs. 

With these insights, we identify two top-performing models: GTs with BI-MPNN layers (effective for small graphs in the HLS and AMP datasets) and the BI-Graph Isomorphism Network (GIN)~\cite{xu2018powerful}, both enhanced by stable PEs. These models outperform all baselines originally designed by hardware experts for corresponding tasks, across all 13 tasks. 
Notably, this work is the first to consider GTs with BI-MPNN layers and using stable PEs in DGRL, so the above two models have novel architectures essentially derived from our benchmarking effort. 

Furthermore, recognizing that hardware design often encounters out-of-distribution (OOD) data in production (e.g., from synthetic to real-world~\cite{wu2022high}, before and after technology mapping~\cite{wu2023gamora}, inference on different RISC-V CPUs~\cite{he2021graph}), for each dataset we evaluate the methods data with distribution shift to simulate potential OOD challenges. We observe that while ML models perform reasonably well on tasks (8 of 13) with diverse graph structures in the training dataset, they generally suffer from OOD generalization issues on the remaining tasks. This finding highlights the urgent need for future research to focus on improving the OOD generalization capabilities of DGRL models.

Lastly, our benchmark is implemented with a modular and user-friendly codebase, allowing hardware practitioners to evaluate all 21 DGRL models for their tasks with data in a PyG-compatible format~\cite{Fey/Lenssen/2019}, and allowing ML researchers to  advance DGRL methods using the collected hardware design tasks. 

\begin{table}[t]
\vspace{-0.2cm}
\resizebox{\textwidth}{!}{\begin{tabular}{@{}ccccccc@{}}
\toprule
\multicolumn{2}{c}{} & \begin{tabular}[c]{@{}c@{}}High-level Synthesis \\ (HLS)~\cite{wu2022high}\end{tabular} & \begin{tabular}[c]{@{}c@{}}Symbolic Reasoning \\ (SR)~\cite{wu2023gamora}\end{tabular} & \begin{tabular}[c]{@{}c@{}}Pre-routing Timing Prediction \\ (Time)~\cite{guo2022timing}\end{tabular} & \begin{tabular}[c]{@{}c@{}}Computational Graph \\ (CG)~\cite{zhang2021nn}\end{tabular} & \begin{tabular}[c]{@{}c@{}}Operational Amplifiers \\ (AMP)~\cite{dong2023cktgnn}\end{tabular} \\ \midrule
\multicolumn{2}{c}{Type} & digital & digital & digital & digital & analog \\ \midrule
\multicolumn{2}{c}{Level} & graph & node & node & graph & graph \\ \midrule
\multicolumn{2}{c}{Target} & regression & classification & regression & regression & regression \\ \midrule
\multicolumn{2}{c}{Task} & LUT, DSP, CP & \begin{tabular}[c]{@{}c@{}}node shared by MAJ and XOR,\\root node of an adder\end{tabular} & \begin{tabular}[c]{@{}c@{}}hold slack,\\setup slack\end{tabular} & CPU/GPU630/GPU640 & gain, PM, BW \\ \midrule
\multicolumn{2}{c}{Evaluation Metric} & mse, r2 & \begin{tabular}[c]{@{}c@{}}accuracy,  f1\\ recall, precision \end{tabular} & mse, r2 & rmse, acc5, acc10 & mse, rmse \\ \midrule
\multicolumn{2}{c}{In-Distribution} & CDFG & 24-bit & graph structure & network structure & stage3 \\ \midrule
\multicolumn{2}{c}{Out-of-Distribution} & DFG & 32, 36, 48- bit & graph structure & network structure & stage2 \\ \midrule
\multicolumn{2}{c}{\# Training Graph} & 16570 - 16570 & 1 - 1 & 7 - 7 & 5* - 10000 & 7223-7223 \\ \midrule
\multirow{2}{*}{\#Train Nodes} & average & 95 & 4440 & 29839 & 218 & 9 \\
 & max & 474 & 4440 & 58676 & 430 & 16 \\ \midrule
\multirow{2}{*}{\# Train Edges} & average & 123 & 10348 & 41268 & 240 & 15 \\
 & max & 636 & 10348 & 83225 & 487 & 36 \\ \bottomrule
\end{tabular}}
\captionsetup{font=small}
\caption{Statistics of selected datasets. In row `\# Training graph', we report `\# Graph Structures - \# Samples'. *: in CG, there are only five unique CNN designs, yet the structure of graphs within each design may vary slightly.}
\label{tab:datasets}
\vspace{-0.4cm}
\end{table}

\section{Related Work}
\label{sec:related work}




\textbf{Graph Representation Learning as Powerful Surrogate Models.}
ML-based surrogate models have been widely adopted in scientific fields~\cite{olivier2021bayesian, zuo2021accelerating} 
and recently extended in hardware design.
While graph-learning-based surrogate models for hardware design have already demonstrated effectiveness~\cite{wang2022unsupervised, ustun2020accurate,wu2022high,wu2021ironman,bai2023towards,ren2020paragraph,zhang2019circuit,li2020customized,ma2019high,vasudevan2021learning,bucher2022appgnn,gnn_re,lu2023eco}, several aspects warrant further investigation.
First, existing studies often rely on task-specific heuristics to encode circuit structural information~\cite{ma2019high,ren2020paragraph,gnn_re,bucher2022appgnn,vasudevan2021learning,mirhoseini2021graph}, hindering the migration of model-design insights from one task to an even closely related task.
Second, the majority of these studies conduct message passing of GNNs along edge directions, with few considering BI implementation~\cite{he2021graph,guo2022timing}, and there is an absence of a comparative analysis of different DGRL approaches.
Third, the designed models are often trained and tested within similar data distributions~\cite{gnn_re,zhao2022graph,he2021graph}, lacking systematic OOD evaluation for new or more complicated designs.
Hence, it is imperative to establish a comprehensive benchmark to compare different DGRL approaches for hardware design tasks.

\textbf{Methods for DGRL.} NN architectures for DGRL can be classified into three types: spatial GNNs, spectral GNNs, and transformers. Spatial GNNs use graph topology as inductive bias, some employ bidirected message passing for regular directed graphs~\cite{jaume2019edgnn, wen2020neural, kollias2022directed, rossi2024edge}, others use asynchronous message passing exclusively designed for directed acyclic graphs (DAGs)~\cite{zhang2019d, dong2022pace, thost2020directed}. 
Spectral GNNs generalize the ideas of Fourier transform and corresponding spectral convolution from undirected to directed graphs~\cite{furutani2020graph, he2022msgnn, fiorini2023sigmanet, zhang2021magnet, singh2016graph, ma2019spectral,monti2018motifnet,tong2020directed,koke2023holonets}; Transformers with attention mechanism reply on designing direction-aware PEs to capture directed graph topology. This benchmark is the first to consider combining transformers with MPNN layers for DGRL, extending the ideas in~\cite{rampavsek2022recipe}. Regarding the choices of PEs, most studies are on undirected graphs~\cite{wang2022equivariant,huang2023stability,lim2022sign,dwivedi2022graph}. For directed graphs, the potential PEs are Laplacian eigenvectors of the undirected graphs by symmetrizing the original directed ones~\cite{dwivedi2020benchmarking}, singular vectors of adjacency matrices~\cite{hussain2022global} and the eigenvectors of Magnetic Laplacians~\cite{shubin1994discrete,fanuel2017magnetic,fanuel2018magnetic,geisler2023transformers}. No previous investigate benefit for DGRL from stably incorporating PE~\cite{wang2022equivariant,huang2023stability}, and we are the first to consider stable PEs for DGRL.


\textbf{Existing Relevant Benchmarks.} 
Dwivedi et al.~\cite{dwivedi2022LRGB} benchmark long-range reasoning of GNNs on undirected graphs; 
PyGSD~\cite{he2024pytorch} benchmarks signed and directed graphs, while focusing on social or financial networks. 
We also compare all the methods for directed unsigned graphs in PyGSD and notice that the SOTA spectral method therein - MagNet~\cite{zhang2021magnet} still works well on node-level tasks on a single graph (SR), which shares some similar insights. 
The hardware community has released graph-structured datasets from various development stages to assist surrogate model development, 
including but not limited to NN workload performance~\cite{zhang2021nn,phothilimthana2024tpugraphs}, CPU throughput~\cite{bhive,sykora2022granite,mendis2019ithemal}, resource and timing in HLS~\cite{wu2022high,bai2023towards}, design quality in logic synthesis~\cite{chowdhury2021openabc}, design rule checking in physical synthesis~\cite{guo2022timing,circuitnet_1,2024circuitnet,chhabria2024openroad}, and hardware security~\cite{yu2021hw2vec}.
In addition to datasets, ProGraML~\cite{cummins2021a} introduces a graph-based representation of programs derived from compiler IRs (e.g., LLVM/XLA IRs) 
for program synthesis and compiler optimization. 
Very recently, Google launched  TPUgraph for predicting the runtime of ML models based on their computational graphs on TPUs~\cite{phothilimthana2024tpugraphs}.
Our CG dataset includes computational graphs of ML models, specifically on edge devices. 

\vspace{-2mm}
\section{Datasets and Tasks}
\vspace{-1mm}
This section introduces the five datasets with thirteen tasks used in this benchmark.  
The datasets cover both digital and analog hardware, considering different circuit abstraction levels, as illustrated in Fig.~\ref{fig:representativeness_datasets}. 
Table~\ref{tab:datasets} displays the statistics of each dataset. 
Next, we briefly introduce the five datasets, with details provided in Appendix.~\ref{app:dataset_selection_detail}. 
Although these datasets are generated by existing studies, we offer modular pre-processing interfaces to make them compatible with PyTorch Geometric and user-friendly for integration with DGRL methods. 

\textbf{High-Level Synthesis (HLS)~\cite{wu2022high}:} 
The HLS dataset collects IR graphs of C/C++ code after front-end compilation~\cite{alfred2007compilers}, and provides post-implementation performance metrics on FPGA devices as labels for each graph, which are obtained after hours of synthesis with Vitis~\cite{vitishls} and implementation with Vivado~\cite{vivado}. 
The labels to predict include resource usage, (i.e., look-up table (LUT) and digital signal processor (DSP)), and the critical path timing (CP). See Appendix.~\ref{app:hls_description} for graph input details. 

\vspace{-0.1cm}
\textit{Significance:}
The HLS dataset is crucial for testing NNs' ability to accurately predict post-implementation metrics 
to accelerate design evaluation in the stage of HLS. 

\vspace{-0.1cm}
\textit{OOD Evaluation:} For training and ID testing, we use control data flow graphs (CDFG) that integrate control conditions with data dependencies, derived from general C/C++ code; As to OOD cases, we use data flow graphs (DFG) derived from basic blocks, leading to distribution shifts.

\textbf{Symbolic Reasoning (SR)~\cite{wu2023gamora}:} 
The SR dataset collects bit-blasted Boolean networks (BNs) (unstructured gate-level netlists), with node labels annotating high-level abstractions on local graph structures, e.g., XOR functions, majority (MAJ) functions, and adders, generated by the logic synthesis tool ABC~\cite{brayton2010abc}. Each graph supports two tasks: root nodes of adders, and nodes shared by XOR and MAJ functions. See Appendix.~\ref{app:sr_description} for detailed input encoding and label explanation. 

\vspace{-0.1cm}
\textit{Significance:} Reasoning high-level abstractions from BNs has wide applications in improving functional verification efficiency~\cite{ciesielski2019understanding} and malicious logic identification~\cite{mahzoon2019revsca}.
GNN surrogate models are anticipated to replace the conventional structural hashing and functional propagation~\cite{li2013wordrev,subramanyan2013reverse} and boost the scalability with significant speedup.
For graph ML, due to significant variation in the size of gate-level netlists under different bit widths, SR is an ideal real-world application to evaluate whether GNN designs can maintain performance amidst the shifts in graph scale.

\vspace{-0.1cm}
\textit{OOD Evaluation:} We use a $24$-bit graph ($4440$ nodes) for training, and $32, 36, 48$-bit graphs (up to $18096$ nodes) for ID testing, derived from carry-save-array multipliers before technology mapping. 
OOD testing data are multipliers after ASAP $7$nm technology mapping~\cite{xu2017standard} with the same bits.

\textbf{Pre-routing Timing Prediction (TIME)~\cite{guo2022timing}:} 
The TIME dataset collects real-world circuits with OpenROAD~\cite{openroad} on SkyWater $130$nm technology~\cite{skywater}.
The goal is to predict slack values at timing endpoints for each circuit design by using pre-routing information.
Two tasks are considered: hold slack and setup slack.
Details are provided in Appendix.~\ref{app:time_description}.

\vspace{-0.1cm}
\textit{Significance:} In physical synthesis, timing-driven placement demands accurate timing information, which is only available after routing.
Repetitive routing and static timing analysis provide accurate timing but are prohibitively expensive.
ML models that precisely learn routing behaviors and timing computation flows are highly expected to improve the efficiency of placement and routing.

\vspace{-0.1cm}
\textit{OOD Evaluation:} We divide ID-OOD based on the difference in graph structures (e.g. `blabla' and `xtea' are different circuit designs, allocated into ID or OOD groups). See details in Appendix.~\ref{app:time_distirbution_definition}.

\textbf{Computational Graph (CG)~\cite{zhang2021nn}:} 
The CG dataset consists of computational graphs of convolutional neural networks (CNNs) with inference latency on edge devices (i.e., Cortex A76 CPU, Adreno 630 GPU, Adreno 640 GPU) as labels. 
The CNNs have different operator types or configurations, either manually designed or found by neural architecture search (NAS).
Details are in Appendix.~\ref{app:cg_description}.

\vspace{-0.1cm}
\textit{Significance:} Accurately measuring the inference latency of DNNs is essential for high-performance deployment on hardware platforms or efficient NAS~\cite{ren2021comprehensive, shi2022nasa}, which however is often costly.
ML-based predictors offer the potential for design exploration and scaling up to large-scale hardware platforms.


\vspace{-0.1cm}
\textit{OOD Evaluation:} We split ID-OOD with different graph structures. (e.g. `DenseNets' and `ResNets' are CNNs with different structures, allocated into different groups). See Appendix.~\ref{app:cg_distribution_definition} for details.

\textbf{Operational Amplifiers (AMP)~\cite{dong2023cktgnn}:} 
AMP dataset contains $10,000$ distinct 2- or 3-stage operational amplifiers (Op-Amps). Circuit specifications (i.e. DC gain, phase margin (PM), and bandwidth (BW)) as labels are extracted after simulation with Cadence Spectre~\cite{spectre}.
Details are in Appendix.~\ref{app:amp_description}.

\vspace{-0.1cm}
\textit{Significance:} Analog circuit design is less automated and requires more manual effort compared to its digital counterpart. Mainstream approaches such as SPICE-based circuit synthesis and simulation~\cite{vladimirescu1994spice}, are computationally expensive and time-consuming.
If ML algorithms can approximate the functional behavior and provide accurate estimates of circuit specifications, they may significantly reduce design time by minimizing reliance on circuit simulation~\cite{afacan2021machine}. 

\vspace{-0.1cm}
\textit{OOD Evaluation:} 
For training and ID testing, we use 3-stage Op-Amps, which have three single-stage Op-Amps in the main feed-forward path).
For OOD evaluation, we use 2-stage Op-Amps.

\textbf{Extensions} Although the datasets cover different levels of circuit abstraction, there are additional tasks in hardware design worth exploration with DGRL surrogates, as reviewed in Section~\ref{sec:related work}. Our modular benchmark framework allows for easy extension to accommodate new datasets.


\section{Benchmark Design}

\subsection{Design Space for Directed Graph Representation Learning}

\begin{figure}[t] 
    \centering
    \begin{minipage}{0.55\textwidth} 
        \centering
        \includegraphics[width=\linewidth]{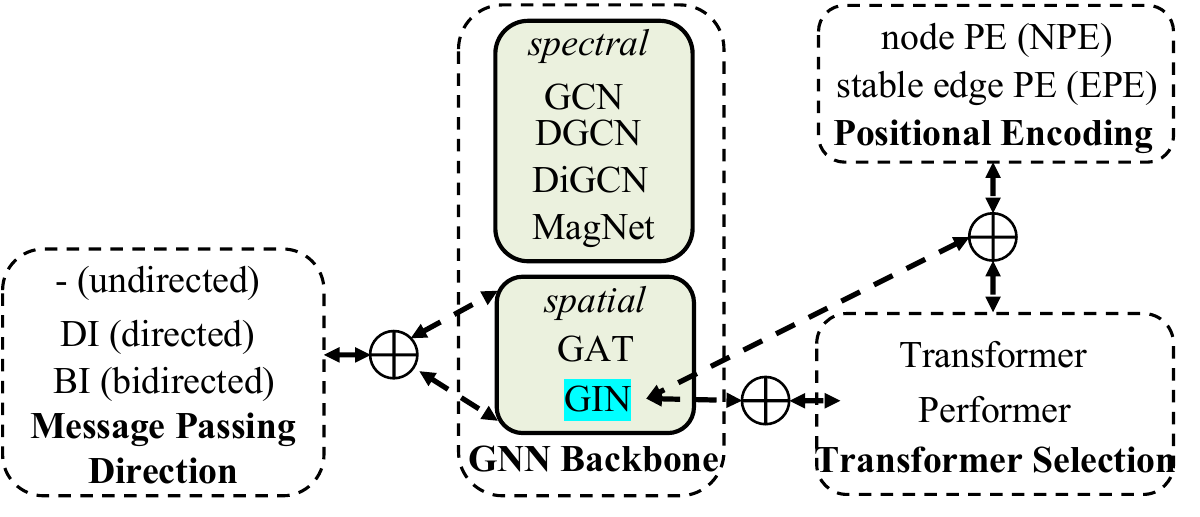} 
        \caption{The benchmark considers 21 combinations of message passing direction, GNN backbone, transformer selection and PE incorporation, covers 10 existing SOTA methods from graph ML community and discovers 2 novel top-performing models (see Table.~\ref{tab:method_eixsting_proposed}).}
        \label{fig:method_combination}
    \end{minipage}
    \hfill 
    \begin{minipage}{0.43\textwidth} 
        \centering
        \resizebox{\textwidth}{!}{\begin{tabular}{@{}ccc@{}}
\toprule
Method & type & \begin{tabular}[c]{@{}c@{}}layer-wise\\ complexity\end{tabular} \\ \midrule
GCN~\cite{kipf2016semi} & spectral & $O(|E|)$ \\
MagNet~\cite{zhang2021magnet} & spectral & $O(|E|)$ \\
DGCN~\cite{tong2020directed} & spectral & $O(|E|)$ \\
DiGCN~\cite{tong2020digraph} & spectral & $O(|E|)$ \\ \midrule
GAT~\cite{velivckovic2017graph} & spatial & $O(|E|)$ \\
GIN(E)~\cite{xu2018powerful} & spatial & $O(|E|)$ \\
EDGNN~\cite{jaume2019edgnn} & spatial & $O(|E|)$ \\ \midrule
GPS-T~\cite{rampavsek2022recipe} & spatial+transformer & $O(|V|^2 + |E|)$ \\
GPS-P~\cite{choromanski2020rethinking} & spatial+transformer & $O(|V|+|E|)$ \\
TmD~\cite{geisler2023transformers} & transformer & $O(|V|^2)$ \\ \midrule
BI-GIN(E)+EPE(new) & spatial & $O(|E|)$ \\
BI-GPS-T+EPE(new) & spatial+transformer & $O(|V|^2 + |E|)$ \\ \bottomrule
\end{tabular}}
\vspace{-0.cm}
\captionof{table}{Existing methods and two top-performing methods highlighted at bottom.}
\label{tab:method_eixsting_proposed}
    \end{minipage}
\vspace{-0.5cm}
\end{figure}

In this section, we introduce the DGRL methods evaluated in this benchmark. Our evaluation focuses on four design modules involving GNN backbones, message passing directions, transformer selection, and PE incorporation, illustrated in Fig.~\ref{fig:method_combination}. Different GNN backbones and transformer adoptions cover 10 methods in total with references in Tab.~\ref{tab:method_eixsting_proposed}. 
We also consider their combinations with different message-passing directions and various ways to use PEs, which overall gives 21 DGRL methods. 

For GNNs, we consider 4 spectral methods, namely GCN~\cite{kipf2016semi}, DGCN~\cite{tong2020directed}, DiGCN~\cite{tong2020digraph} and MagNet~\cite{zhang2021magnet}, where the latter three are \emph{SOTA spectral GNNs} specifically designed for DGRL~\cite{he2024pytorch}; 
For spatial GNNs, we take GIN~\cite{xu2018powerful} and Graph Attention Network (GAT)~\cite{velivckovic2017graph}, which are the most commonly used MPNN backbones for undirected graphs. We evaluate the combination of GCN, GIN and GAT with three different message-passing directions: a) `undirected'(-) treats directed graphs as undirected, using the same NN parameters to perform message-passing along both forward and reverse edge directions; b) `directed'(DI) only passes messages exclusively along the forward edge directions; c) `bidirected'(BI) performs message passing in both forward and reverse directions with distinct parameters for either direction. The other GNNs (DGCN, DiGCN and MagNet)  adopt spectral convolution that inherently considers edge directions. The combination of `BI' with spatial GNN layers gives \emph{the state-of-the-art spatial GNNs} for DGRL, i.e., EDGNN~\cite{jaume2019edgnn}. 

For GTs,  
 we adopt the eigenvectors of the graph Magnetic Laplacian (MagLAP) matrix as the PEs of nodes~\cite{furutani2020graph, shubin1994discrete}, as they are directional-aware. The MagLap matrix $\bm{L}_q$ is a complex Hermitian matrix with parameter $q\in[0, 1)$ named potential, which is treated as a hyper-parameter in our experiments. Note that when $q=0$, MagLap degenerates to the symmetric Laplacian matrix $\bm{L}_{0}$  as a special case. See Appendix~\ref{app:pe} for a brief review of MagLap. The GT with the MagLap PEs attached to node features gives \emph{the SOTA GT model} for DGRL, named TmD for brevity, proposed in~\cite{geisler2023transformers}. GPS~\cite{rampavsek2022recipe} is a GT model with MPNN layers~\cite{hamilton2017inductive,gilmer2017neural} interleaving with transformer layers~\cite{vaswani2017attention}, originally proposed for undirected graphs. We extend GPS to directed graphs by using MagLap PEs for transformer layers and DI/BI message passing in its MPNN layers. Hence, GPS is also an extension of TmD by incorporating MPNN layers. As transformers may not scale well on large graphs, we evaluate vanilla transformer layers and their lower-rank approximation Performer~\cite{kreuzer2021rethinking} for efficient computation, named as GPS-T and GPS-P, respectively.



\subsection{Stable Direction-aware Positional Encodings}
\vspace{-1mm}
\label{sec:pe}

\begin{wraptable}{r}{0.6\textwidth}
\vspace{-0.3cm}
\captionsetup{font=footnotesize}
\centering
\resizebox{0.6\textwidth}{!}{\begin{tabular}{@{}ccc@{}}
\toprule
 $  \textbf{NPE}=[\text{Re}\{\bm{V}_q\}, \text{Im}\{\bm{V}_q\}]$           \\ \midrule
  $ \textbf{EPE}= \rho(\text{Re}\{\bm{V_q}\text{diag}(\kappa_1(\lambda))\bm{V_q}^{\dagger}\}, ..., \text{Re}\{\bm{V_q}\text{diag}(\kappa_m(\lambda))\bm{V_q}^{\dagger}\},$   \\ 
  \hspace{30pt}$\text{Im}\{\bm{V_q}\text{diag}(\kappa_1(\lambda))\bm{V_q}^{\dagger}\}, ..., \text{Im}\{\bm{V_q}\text{diag}(\kappa_m(\lambda))\bm{V_q}^{\dagger}\})$  \\\bottomrule
\end{tabular}}
\vspace{-0.cm}
\caption{Functions to obtain PEs. NPE directly concatenates the eigenvectors to node features. In contrast, before concatenating PE to the edge features, EPE employs the permutation equivariant functions $ \kappa : \mathbb{R}^{d} \rightarrow \mathbb{R}^d$ w.r.t. eigenvalue permutations and permutation equivariant function $\rho: \mathbb{R}^{|V|\times |V| \times 2m} \rightarrow \mathbb{R}^{|V| \times |V| \times d}$ to stably process the eigenvectors and eigenvalues, respectively.}
\label{tab:node_edge_PE}
\vspace{-0.4cm}
\end{wraptable}
Recent studies on undirected graphs have demonstrated that models by naively attaching PEs to node features may suffer from an issue of instability because small changes in the graph structure may cause big changes in PEs~\cite{wang2022equivariant,huang2023stability,lim2022sign}. We name this way of using PEs as node-PE (NPE). The instability provably leads to undesired OOD generalization~\cite{huang2023stability}. We think this is also true for directed graphs and indeed observe the subpar model performance with NPE.

Therefore, besides NPE, we also consider a stable way of incorporating PEs for DGRL, namely `edge PE' (EPE), inspired by~\cite{wang2022equivariant}. EPE was originally proposed for the undirected graph case. Specifically, we use the smallest $d$ eigenvalues $\lambda_q \in \mathbb{R}^{d}$ and their corresponding eigenvectors $\bm{V}_q \in \mathbb{C}^{|V| \times d}$ from $\bm{L}_q$.
Then, we follow the equation in Table~\ref{tab:node_edge_PE} to compute $\textbf{EPE}\in \mathbb{R}^{|V| \times |V| \times d}$. Then, in GTs, $\textbf{EPE}_{u,v}$ is further added to the attention weight between nodes $u$ and $v$ as a bias term at each attention layer. 

We note that PEs can also be used in more than GTs, to improve the expressive power of GNNs ~\cite{li2020distance,ying2021transformers,lim2022sign,huang2023stability}. We leverage this idea and enhance the GNN models for directed graphs with PEs. Specifically, for the GNNs NPE will use $\textbf{NPE}_v$ as extra node features of node $v$ while EPE will use $\textbf{EPE}_{u,v}$ as extra edge features of edge $uv$ if $uv$ is an edge. 

The incorporation with EPE helps discover a novel GT model for  directed graphs, i.e., GT with BI-MPNN layers enhanced by EPE, abbreviated as BI-GPS+EPE. 
We also make the first attempt to combine GNNs with PEs for directed graphs, which yields the model BI-GIN(E)+EPE. 

\subsection{Hyer-Parameter Space and Tuning}
For each combination of DGRL method in this benchmark, we perform automatic hyper-parameter tuning with RAY~\cite{liaw2018tune} adopting Tree-structured Parzen Estimator (TPE)~\cite{watanabe2023tree}, a state-or-the-art bayesian optimization algorithm. The hyper-parameter space involves searching batch size, learning rate, number of backbone layers, dropout rate in MPNN and MLP layers, hidden dimension, and MLP layer configurations. The detailed hyper-parameter space of each model is shown in Appendix.~\ref{app:hyper-parameter_space}. We auto-tune the hyper-parameters with seed $123$ with $100$ trial budgets and select the configuration with the best validation performance. Then, the selected configuration is used for model training and testing ten times with seeds $0-9$ and the average is reported as the final performance.

\section{Modular Toolbox}
\begin{figure}[h]
\vspace{-0.3cm}
    \centering
    \includegraphics[width = 0.8 \linewidth]{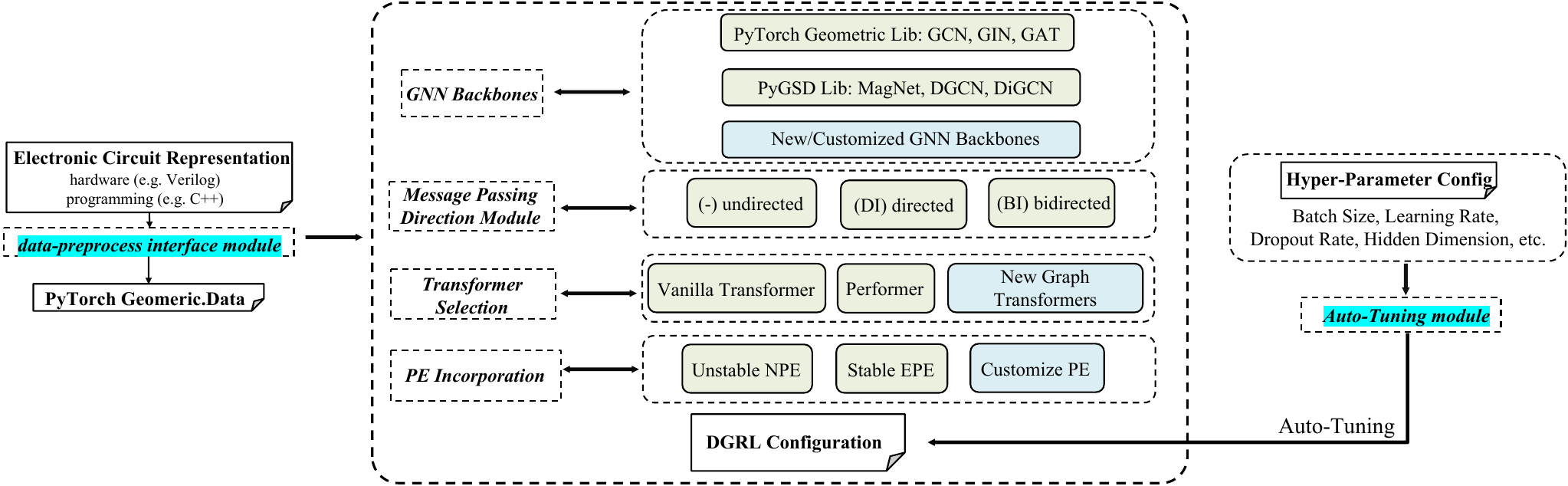}
    \vspace{-0.1cm}
    \caption{Illustration of the directed graph representation learning (DGRL) toolbox.}
    \label{fig:toolbox}
    \vspace{-0.4cm}
\end{figure}
We develop a highly modular toolbox involving designing, auto hyper-parameter tuning, and evaluation for DGRL methods. The framework is shown in Fig.~\ref{fig:toolbox}. The toolbox comes with the 21 DGRL methods, allowing practitioners to evaluate them on any new task with data compatible with PyTorch Geometric (PyG)~\cite{Fey/Lenssen/2019}. This may be used even beyond hardware design applications. Users can also customize new methods. Once the method is configured, auto hyper-parameter tuning can be performed using RAY~\cite{liaw2018tune}. The toolbox also includes the above 5 datasets with 13 tasks that can be used to develop new DGRL models. For details please refer to the official document for this toolbox\url{}.

\section{Experiments}

In this section, we first evaluate DGRL methods combining different  GNN backbones, message passing directions, transformer selection, and PE incorporation, across all 5 datasets and 13 tasks, using in-distribution (ID) and out-of-distribution (OOD) testing data. 

\subsection{Main Results}
The performances of the methods under all evaluation metrics for both in-distribution and out-of-distribution testing across all 13 tasks are reported from Table.~\ref{tab:sr_id} to Table.~\ref{tab:cg_gpu640_ood} in Appendix.~\ref{app:exp_main_result}. We summarize the averaged ranking with respect to all evaluation metrics given a task in Table.~\ref{tab:exp_gnn_backbone_selection}. The details of ranking calculation is in Appendix.~\ref{app:ranking_expression}. The results tell the following insights: 

`Bidirected' (BI) message passing in the MPNN layers significantly boosts the models' performance on three GNN backbones (GCN, GIN, GAT) and one GT backbone (GPS-T): BI-GCN outperforms GCN on 10 out of 13 tasks in both ID and OOD evaluations. Similarly, in ID/OOD evaluations, BI-GIN outperforms GIN in 11/12 out of 13 tasks, BI-GAT outperforms GAT in 11/9 out of 13 tasks and BI-GPS-T outperforms GPS-T in 5/5 out of 6 tasks, respectively. 

As to the models, on datasets with small graphs (HLS and AMP), BI-GPS-T consistently delivers excellent results, achieving top-3 performance in 5 out of 6 tasks on both ID and OOD testing data. BI-GIN also demonstrates competitive performance on these datasets. However, for datasets with larger graphs (SR, CG, and TIME), BI-GPS-T encounters a scalability issue.  BI-GIN secures top-three performance in 6 out of 7 tasks in both ID and OOD testing data. For the `shared' and `root' tasks from the SR dataset and the `CPU' and `GPU630' tasks from the CG dataset, MagNet~\cite{zhang2021magnet} performs best in the ID setting. This is likely because training and testing are conducted on the same graph structures for these specific datasets, reducing the need for significant generalization across different graph structures. This scenario aligns well with the spectral filtering approach used by MagNet. These observations match findings from previous studies on directed networks~\cite{zhang2021magnet, he2024pytorch}. However, MagNet's performance falters in OOD evaluations which ask for the ability to generalize across different graph structures. GPS-P, despite its capability to handle large graphs, delivers only mediocre performance overall. 
\begin{table}[t]
\setlength{\tabcolsep}{1pt}
\resizebox{\textwidth}{!}{\begin{tabular}{@{}cc|ccc|ccc|cc|cc|ccc||ccc|ccc|cc|cc|ccc@{}}
\toprule
\multicolumn{2}{c}{Distribution} & \multicolumn{13}{|c||}{In-Distribution (ID)} & \multicolumn{13}{c}{Out-of-Distribution (OOD)} \\ \midrule
\multicolumn{2}{c}{Dataset} & \multicolumn{3}{|c|}{HLS} & \multicolumn{3}{c|}{AMP} & \multicolumn{2}{c|}{SR} & \multicolumn{2}{c|}{TIME} & \multicolumn{3}{c||}{CG} & \multicolumn{3}{c|}{HLS} & \multicolumn{3}{c|}{AMP} & \multicolumn{2}{c|}{SR} & \multicolumn{2}{c|}{TIME} & \multicolumn{3}{c}{CG} \\ \midrule
\multicolumn{2}{c|}{Task} & DSP & LUT & CP & gain & PM & BW & share & root & hold & setup & CPU & GPU630 & GPU640 & DSP & LUT & CP & gain & PM & BW & share & root & hold & setup & CPU & GPU630 & GPU640 \\ \midrule
 & DGCN & 15.0 & 15.0 & 15.0 & 14.0 & 8.0 & 15.0 & 10.0 & 9.0 & 15.0 & 5.5 & 13.0 & 15.0 & 14.0 & 15.0 & 14.0 & 15.0 & 14.0 & \cellcolor{3}3.0 & 15.0 & 7.5 & 5.0 & 15.0 & 7.0 & 13.3 & 11.7 & 11.2 \\
 & DiGCN & 12.0 & 14.0 & 13.0 & 12.0 & 9.0 & 14.0 & 8.5 & 7.8 & 13.5 & 15.0 & 14.0 & 14.0 & 15.0 & 12.5 & 15.0 & 14.0 & 9.0 & 4.0 & 14.0 & 9.0 & 5.0 & 13.5 & 14.0 & 13.2 & 13.2 & 13.3 \\
 & MagNet & 7.0 & 7.0 & 10.5 & 8.0 & 11.0 & 8.0 & \cellcolor{1}1.8 & \cellcolor{1}2.0 & 11.0 & 11.5 & \cellcolor{1}1.3 & \cellcolor{1}1.3 & 4.7 & 7.0 & 7.0 & 10.5 & \cellcolor{3}3.0 & 12.0 & 8.0 & \cellcolor{3}3.5 & 8.8 & 9.0 & 7.0 & \cellcolor{3}4.2 & 8.2 & 7.3 \\
 & GCN & 14.0 & 12.0 & 14.0 & 15.0 & 13.0 & 12.0 & 13.3 & 13.5 & 9.5 & 14.0 & 15.0 & 12.3 & 11.7 & 12.5 & 10.0 & 12.0 & 14.5 & 14.0 & 11.0 & 14.8 & 14.5 & 7.5 & 10.5 & 12.7 & 12.7 & 11.5 \\
 & DI-GCN & 13.5 & 13.0 & 12.0 & 11.0 & \cellcolor{3}3.0 & 13.0 & 15.0 & 15.0 & 11.0 & 13.0 & 11.0 & 11.3 & 12.0 & 14.0 & 11.0 & 13.0 & 12.0 & 7.0 & 13.0 & 13.5 & 11.8 & 10.0 & 8.0 & 11.2 & 11.5 & 12.2 \\
\multirow{-6}{*}{Spectral} & BI-GCN & 11.0 & 10.5 & 9.0 & 5.0 & 14.0 & 6.0 & 5.5 & 5.3 & 5.0 & 9.0 & 12.3 & 12.3 & 12.3 & 11.0 & 12.5 & 8.0 & \cellcolor{2}2.0 & 13.0 & 5.0 & \cellcolor{2}2.3 & \cellcolor{3}4.8 & \cellcolor{2}3.0 & \cellcolor{2}6.5 & 13.2 & 11.3 & 12.5 \\ \midrule
 & GIN & 6.0 & 5.5 & 8.0 & 7.0 & 6.0 & 10.0 & 10.0 & 11.0 & \cellcolor{1}1.0 & \cellcolor{3}3.0 & 5.0 & \cellcolor{3}3.3 & 8.3 & 6.0 & \cellcolor{3}3.5 & 5.0 & 8.0 & 10.0 & 7.0 & 9.0 & 7.3 & \cellcolor{2}3.0 & 8.5 & 5.2 & \cellcolor{1}4.2 & 4.8 \\
 & DI-GIN & \cellcolor{2}2.5 & 4.0 & 6.5 & 9.0 & 10.0 & 7.0 & 6.5 & \cellcolor{2}4.8 & \cellcolor{3}3.0 & \cellcolor{2}2.0 & 5.7 & 8.0 & \cellcolor{3}3.3 & \cellcolor{1}2.0 & \cellcolor{2}2.5 & 7.0 & 10.0 & 5.0 & 12.0 & 6.3 & 9.0 & 6.5 & 7.0 & \cellcolor{2}3.5 & 5.7 & \cellcolor{2}4.2 \\
 & BI-GIN & \cellcolor{1}1.0 & \cellcolor{1}1.0 & 5.0 & \cellcolor{3}3.0 & 4.0 & \cellcolor{3}3.0 & \cellcolor{2}2.8 & \cellcolor{2}4.8 & \cellcolor{2}2.0 & \cellcolor{1}1.0 & \cellcolor{2}2.3 & 4.7 & \cellcolor{1}1.0 & 4.5 & \cellcolor{1}1.0 & \cellcolor{2}3.0 & 4.0 & 9.0 & 3.0 & \cellcolor{1}1.5 & \cellcolor{1}2.0 & \cellcolor{1}1.0 & 7.5 & \cellcolor{1}2.3 & \cellcolor{3}4.5 & \cellcolor{1}4.0 \\
 & GAT & 8.5 & 9.0 & 6.5 & 6.0 & 15.0 & 5.0 & 13.8 & 13.5 & 10.5 & 11.5 & 9.0 & 9.0 & 8.7 & 9.0 & 9.0 & 5.5 & 7.0 & 15.0 & 6.0 & 12.3 & 10.5 & 10.5 & \cellcolor{1}5.5 & 7.7 & 5.7 & 6.2 \\
 & DI-GAT & 10.0 & 10.5 & 10.5 & 10.0 & 12.0 & 9.0 & 11.8 & 10.0 & 13.5 & 10.0 & 10.0 & 10.0 & 10.0 & 10.0 & 12.5 & 11.0 & 11.0 & 6.0 & 10.0 & 11.3 & 10.0 & 13.5 & 8.5 & 6.2 & 5.7 & 7.3 \\
\multirow{-6}{*}{Spatial} & BI-GAT & 9.0 & 8.0 & \cellcolor{1}1.0 & 4.0 & \cellcolor{2}2.0 & 11.0 & \cellcolor{3}4.0 & 6.3 & 6.5 & 7.5 & 8.0 & 5.3 & 7.0 & 8.0 & 8.0 & \cellcolor{1}1.5 & \cellcolor{1}1.0 & \cellcolor{2}2.0 & 9.0 & 9.8 & 8.5 & 4.5 & \cellcolor{2}6.5 & 6.2 & 10.7 & 10.5 \\ \midrule
 & GPS-T & 4.0 & \cellcolor{3}3.0 & \cellcolor{3}2.5 & 13.0 & 7.0 & \cellcolor{2}2.0 & - - & - - & - - & - - & - - & - - & - - & 5.0 & 6.0 & 9.5 & 13.0 & 8.0 & \cellcolor{1}1.0 & - - & - - & - - & - - & - - & - - & - - \\
 & DI-GPS-T & 5.0 & 5.5 & 4.0 & \cellcolor{1}1.0 & 5.0 & \cellcolor{1}1.0 & - - & - - & - - & - - & - - & - - & - - & \cellcolor{3}3.0 & 5.0 & \cellcolor{1}1.5 & 5.0 & 11.0 & \cellcolor{2}2.0 & - - & - - & - - & - - & - - & - - & - - \\
 & BI-GPS-T & \cellcolor{3}3.0 & \cellcolor{2}2.0 & \cellcolor{2}2.0 & \cellcolor{1}1.0 & \cellcolor{1}1.0 & 4.0 & - - & - - & - - & - - & - - & - - & - - & \cellcolor{2}2.5 & \cellcolor{3}3.5 & \cellcolor{3}4.5 & 6.0 & \cellcolor{1}1.0 & \cellcolor{3}4.0 & - - & - - & - - & - - & - - & - - & - - \\
 & GPS-P & - - & - - & - - & - - & - - & - - & 5.5 & 12.0 & 6.5 & 4.0 & 6.3 & \cellcolor{2}2.0 & 5.3 & - - & - - & - - & - - & - - & - - & 7.8 & 11.3 & 8.0 & 7.5 & 6.2 & \cellcolor{1}4.2 & 6.2 \\
 & DI-GPS-P & - - & - - & - - & - - & - - & - - & 6.5 & \cellcolor{2}4.8 & 4.0 & 7.5 & \cellcolor{3}2.7 & 5.7 & \cellcolor{1}3.0 & - - & - - & - - & - - & - - & - - & 5.8 & 7.5 & 7.5 & 8.0 & 7.5 & 7.0 & 5.8 \\
\multirow{-6}{*}{Transformer} & BI-GPS-P & - - & - - & - - & - - & - - & - - & 7.8 & 7.5 & 8.0 & 5.5 & 4.7 & 6.0 & 4.3 & - - & - - & - - & - - & - - & - - & 6.8 & \cellcolor{2}4.3 & 7.5 & 8.0 & 8.5 & 5.2 & \cellcolor{2}4.2 \\ \bottomrule
\end{tabular}}
\caption{Average ranking ($\downarrow$) of methods across datasets/tasks/metrics on ID and OOD data.}
\vspace{-0.5cm}
\label{tab:exp_gnn_backbone_selection}
\end{table}
\textit{In conclusion, BI-GPS is well-suited for small (around one hundred nodes) directed graphs. For larger graphs, BI-GIN is efficient and performs well. For tasks where the training and testing data share the same graph structures, one may also attempt to adopt MagNet.}

\begin{table}[t]
\setlength{\tabcolsep}{1pt}
\resizebox{\textwidth}{!}{\begin{tabular}{@{}c|ccc|ccc|cc|cc|ccc||ccc|ccc|cc|cc|ccc@{}}
\toprule
Distribution & \multicolumn{13}{c|}{In-Distribution (ID)} & \multicolumn{13}{c}{Out-of-Distribution (OOD)} \\ \midrule
Dataset & \multicolumn{3}{c}{HLS} & \multicolumn{3}{c}{AMP} & \multicolumn{2}{c}{SR} & \multicolumn{2}{c}{TIME} & \multicolumn{3}{c|}{CG} & \multicolumn{3}{c}{HLS} & \multicolumn{3}{c}{AMP} & \multicolumn{2}{c}{SR} & \multicolumn{2}{c}{TIME} & \multicolumn{3}{c}{CG} \\ \midrule
Task & DSP & LUT & CP & gain & PM & BW & share & root & hold & setup & CPU & GPU630 & GPU640 & DSP & LUT & CP & gain & PM & BW & share & root & hold & setup & CPU & GPU630 & GPU640 \\ \midrule
MagNet & 14.5 & 11.0 & 14.5 & 12.0 & 15.0 & 12.0 & \cellcolor{1}2.3 & \cellcolor{1}2.5 & 13.0 & 13.0 & 2.3 & \cellcolor{1}1.7 & 6.7 & 11.0 & 11.0 & 14.5 & \cellcolor{1}3.0 & 16.0 & 12.0 & 5.5 & 10.8 & 11.0 & 16.0 & 5.2 & 9.5 & 8.3 \\ \midrule
BI-GIN(E) & 9.0 & 2.0 & 9.0 & 6.0 & 6.0 & 6.0 & 4.8 & 6.8 & 2.5 & 2.0 & 6.0 & 3.3 & 6.0 & 7.5 & 3.5 & 6.0 & 4.0 & 13.0 & 5.0 & 3.0 & \cellcolor{1}3.0 & 2.5 & \cellcolor{1}7.5 & 4.7 & 5.8 & 4.8 \\
BI-GIN(E)+NPE & 5.0 & 4.0 & 5.0 & 5.0 & 13.0 & 5.0 & 3.0 & 6.8 & 5.5 & 5.0 & 8.3 & 5.0 & 5.3 & 9.0 & 4.0 & \cellcolor{1}1.5 & 7.0 & 8.0 & 7.0 & 2.8 & 3.5 & 5.5 & 12.5 & 6.0 & \cellcolor{1}4.5 & 6.0 \\
BI-GIN(E)+EPE & 5.0 & \cellcolor{1}1.0 & 5.0 & 9.0 & 10.0 & \cellcolor{1}3.0 & 4.0 & 6.8 & \cellcolor{1}2.0 & \cellcolor{1}1.0 & \cellcolor{1}1.0 & 6.7 & \cellcolor{1}1.7 & 7.0 & \cellcolor{1}1.0 & 2.5 & 6.0 & 6.0 & 4.0 & \cellcolor{1}1.5 & \cellcolor{1}3.0 & \cellcolor{1}1.0 & \cellcolor{1}7.5 & \cellcolor{1}3.8 & 5.5 & \cellcolor{1}4.7 \\ \midrule
BI-GPS-T (NPE) & 4.5 & 5.5 & 4.5 & 2.0 & 2.0 & 7.0 & - - & - - & - - & - - & - - & - - & - - & 4.0 & 7.0 & 8.0 & 9.0 & 2.0 & 6.5 & - - & - - & - - & - - & - - & - - & - - \\
BI-GPS-T+EPE & \cellcolor{1}2.5 & 3.0 & \cellcolor{1}2.0 & \cellcolor{1}1.0 & \cellcolor{1}1.0 & 4.0 & - - & - - & - - & - - & - - & - - & - - & \cellcolor{1}1.5 & 1.5 & 3.5 & 5.5 & \cellcolor{1}1.0 & \cellcolor{1}1.0 & - - & - - & - - & - - & - - & - - & - - \\ \bottomrule
\end{tabular}}
\caption{Comparison of competitive methods involving NPE and EPE. The ranking ($\downarrow$) is based on all the 18 methods in Table~\ref{tab:exp_gnn_backbone_selection} plus BI-GIN(E)+NPE, BI-GIN(E)+EPE and BI-GPS-T+EPE.}
\vspace{-0.7cm}
\label{tab:ne}
\end{table}

\begin{table}[t]
\setlength{\tabcolsep}{1pt}
\vspace{-0.3cm}
\resizebox{\textwidth}{!}{\begin{tabular}{@{}c|ccc|ccc|c|ccc|c@{}}
\toprule
\begin{tabular}[c]{@{}c@{}}dataset\\ (baseline's name)\end{tabular} & \multicolumn{3}{c|}{\begin{tabular}[c]{@{}c@{}}AMP~\cite{dong2023cktgnn} \\ (CKTGNN)\end{tabular}} & \multicolumn{3}{c|}{\begin{tabular}[c]{@{}c@{}}HLS~\cite{wu2022high} \\ (Hierarchical GNN)\end{tabular}} & \begin{tabular}[c]{@{}c@{}}SR~\cite{wu2023gamora}\\ (GAMORA)\end{tabular} & \multicolumn{3}{c|}{\begin{tabular}[c]{@{}c@{}}CG~\cite{zhang2021nn}\\ (nn-meter)\end{tabular}} & \begin{tabular}[c]{@{}c@{}}TIME~\cite{guo2022timing} \\ (Timer-GNN)\end{tabular} \\ \midrule
task & gain & PM & BW & dsp & lut & cp & shared & \multicolumn{3}{c|}{cpu (average)} & hold \\ \midrule
metric & rmse$\downarrow$ & rmse$\downarrow$ & rmse$\downarrow$ & mse$\downarrow$ & mse$\downarrow$ & mse$\downarrow$ & accuracy$\uparrow$ & rmse$\downarrow$ & acc5$\uparrow$ & acc10$\uparrow$ & r2$\uparrow$ \\ \midrule
Baseline & 0.52 & 1.15 & 4.47 & 3.94 & 2.45 & 0.88 & 0.99 & 3.20 & 0.80 & 0.99 & 0.97 \\
BI-GINE+EPE & 0.51±0.07 & 1.14±0.00 & 4.20±0.13 & 2.13±0.08 & 1.73±0.10 & 0.61±0.02 & 0.99±0.00 & 2.79±0.14 & 0.86±0.02 & 0.99±0.01 & 0.99±0.00 \\
BI-GPS-T+EPE & 0.34±0.08 & 1.15±0.00 & 3.79±0.11 & 2.13±0.15 & 1.96±0.13 & 0.60±0.01 & - - & - - & - - & - - & - - \\ \bottomrule
\end{tabular}}
\caption{Comparison of BI-GIN+EPE and BI-GPS-T+EPE with baselines specific for each dataset.}
\vspace{-0.8cm}
\label{tab:compare_baselines}
\end{table}
\textbf{Comparing PE-enhanced methods:} We further investigate the impact of different ways of using PEs. We combine NPE or EPE with the top-performing models from the previous section and evaluate BI-GIN+NPE, BI-GIN+EPE, and BI-GPS+EPE. Note that BI-GPS already utilizes NPE. We have chosen not to consider adding PE to MagNet because MagNet only accepts 1-dimensional edge weights, limiting its ability to leverage EPE. We provide a summary of the performance data from Table \ref{tab:amp_id_ne} to Table \ref{tab:cg_ood_ne} in Appendix \ref{app:result_npe_epe} and report the average rankings of the methods for each task. All 18 methods in Table~\ref{tab:exp_gnn_backbone_selection}, along with the 3 new combinations, are included in the ranking. We detail the results of the most competitive methods in Table \ref{tab:ne}. For BI-GIN, EPE enhances its performance on 10 out of 13 tasks in the in-distribution (ID) testing data and 11 tasks in the out-of-distribution (OOD) testing data. Conversely, NPE only improves the performance of BI-GIN on 7 tasks in the ID testing and 4 tasks in the OOD testing and performs unstable for the rest tasks. Notably, EPE-enhanced BI-GIN surpasses MagNet on the CPU task in the CG dataset. For BI-GPS-T, EPE improves its performance on all 6 tasks in both ID and OOD testing, while NPE does not yield substantial improvements. This observation contrasts with previous work \cite{rampavsek2022recipe} on undirected graphs for molecular property prediction. \textit{In conclusion, we find that incorporating PEs in a stable way as EPE significantly boosts the performance of different models across the selected tasks and datasets.}

\subsection{Summary: The Recipe for DGRL}
\vspace{-0.1cm}

Through benchmarking various combinations within the design space, we have formulated a design recipe for DGRL methods tailored for encoding hardware data:
\textit{The use of 'bidirected' (BI) message passing and stable positional encodings (PE) can significantly enhance model performance. Therefore, we recommend BI-GPS-T+EPE for encoding small graphs and BI-GIN+EPE for large graphs.}





We further compare the two models' performance with the baseline methods proposed by hardware design practitioners specifically for the corresponding tasks in the original papers. Results are shown in Table.~\ref{tab:compare_baselines}. The comparison focuses on ID evaluation as for most of the tasks, the original studies did not even report OOD evaluations. We follow the same data split as baseline methods for fair comparison (see the details in Appendix~\ref{app:data-split}).
BI-GIN+EPE achieves results comparable to, or better than, the baseline methods. BI-GPS+EPE achieves even better performance than BI-GIN+EPE for small graphs. Note that the baseline methods for certain tasks may incorporate domain-specific expert knowledge and additional data processing. For example, CKTGNN~\cite{dong2023cktgnn} for the AMP dataset modifies the graph structures into DAGs and employs an asynchronized message passing to mimic the signal flow in these amplifiers; `timer-GNN'~\cite{guo2022timing} is tailored for the TIME dataset to mimic the transmission rules of clock signals and designs a non-linear delay model (NLDM) along with a novel module `cell library'. Such domain knowledge may 
further enhance BI-GPS+EPE and BI-GIN+EPE for these specific tasks, 
which is left for future research.

\textbf{Discussion on OOD Evaluation:} Despite BI-GPS-T+EPE and BI-GIN+EPE outperforming other methods in OOD testing across all tasks, we cannot yet conclude that these methods are sufficiently effective for practical OOD usage. \textit{In fact, making accurate predictions with OOD data in hardware design remains a significant challenge.} When the graph structures in training sets are sufficiently diverse, such as in datasets with a large number of small graphs (e.g., AMP, HLS) or those with abundant local structures (e.g., SR), BI-GIN+EPE and BI-GPS-T+EPE tend to maintain reasonably good performance on OOD data. However, OOD generalization becomes challenging when the diversity of graph structures in the training set is limited. For instance, in the TIME dataset, which has a limited variety of graph structures for training and OOD testing data with entirely different graph structures, both BI-GIN+EPE and BI-GPS-T+EPE perform worse than timer-GNN \cite{guo2022timing}, which integrates the knowledge of the physical structure of circuits (as shown in Table \ref{tab:time_ood}). We identify ensuring OOD performance, especially when training sets lack sufficiently diversified graph structures, as a key direction for future DGRL research.

\vspace{-0.3cm}



\section{Conclusions and Limitations}
\vspace{-0.1cm}
Through benchmarking 21 methods on in-distribution and out-of-distribution test sets across 13 tasks and 5 datasets within the hardware design loop, we find bidirected (BI) message passing neural networks can substantially improve the performance of both Graph Transformer (GT) encoders that incorporate MPNN layers and pure GNN encoders. Positional Encodings (PEs), particularly when used stably, can broadly enhance the performance of both GTs and GNNs. With these insights, we identify two top-performing models: BI-GPS-T+EPE and BI-GIN+EPE, both of which outperform the baseline models originally proposed for the corresponding tasks. 

\textbf{Limitations}: Although the benchmark covers multiple stages in hardware design loop, there are other tasks \cite{mendis2019ithemal, sykora2022granite, 2024circuitnet, circuitnet_1, bhive, gnn_re, zhang2019circuit} that could be included in this benchmark as DGRL tasks. Given technological advancements  and the diversity of design tools, ensuring OOD performance 
remains an urgent open problem in hardware design. Future research may involve high-quality data collection \cite{jain2020overview, gupta2021data, wu2021demonstration, whang2023data, wu2020zeroer} or the development of OOD-aware DGRL methods \cite{liu2023structural, shi2024graph, liu2024pairwise, liu2024beyond}.

\newpage

\bibliographystyle{plain}
\bibliography{reference}

\appendix

\newpage

\section{More Related Work}\label{app:related}
In this section, we review the extensive previous studies that use ML-based surrogate models. 

ML-based surrogate models have been widely used in hardware system design, such as predicting energy/power consumption, latency, throughput, or reliability on CPUs~\cite{lo2015prediction,zheng2016accurate,mishra2018caloree,mendis2019ithemal,lin2019design,bhive,sykora2022granite}, GPUs~\cite{jia2012stargazer,baldini2014predicting,o2017halwpe,pattnaik2016scheduling,chen2018learning,li2020adatune}, tensor processing units (TPUs)~\cite{kaufman2021learned,phothilimthana2024tpugraphs}, and data centers~\cite{murray2005machine,mahdisoltani2017proactive,xu2018improving}.
Similar trends are observed in quickly estimating quality-of-results of circuit designs in EDA flows, spanning high-level synthesis (HLS)~\cite{wang2022unsupervised, zhao2019machine,makrani2019pyramid,ustun2020accurate,lin2020hl,wu2022high, wu2021ironman,bai2023towards, ye2024hida}, logic synthesis~\cite{yu2018developing,yu2020decision,wu2022lostin,wu2023gamora,zhao2022graph}, physical synthesis~\cite{tabrizi2018machine,liang2020drc,chen2020pros,mirhoseini2021graph,guo2022timing,esmaeilzadeh2023open,lu2024gan,lu2020vlsi}, analog circuit designs~\cite{wang2020gcn,zhang2019circuit,ren2020paragraph,shook2020mlparest,li2020customized,dong2023cktgnn}, and design verification~\cite{ma2019high,hughes2019optimizing,shibu2021verlpy,vasudevan2021learning,wu2024survey}. 
\textit{As circuits can naturally be represented as directed graphs, the adoption of GNN-based surrogate models is increasingly prominent.}
We discuss several examples for each of the aforementioned tasks as follows.

In CPU throughput estimation, Granite~\cite{sykora2022granite} adopts a GNN model to predict basic block throughput on CPUs. Basic blocks are represented as graphs to capture the semantic relationships between instructions and registers. A GNN model is then trained to learn expressive embeddings for each basic block, followed by a decoder network to predict the throughput.

In HLS, many studies leverage the IR graphs generated by HLS front-ends.
Ustun et al.~\cite{ustun2020accurate} employs GNNs to predict the mapping from arithmetic operations in IR graphs to different resources on FPGAs. 
IronMan~\cite{wu2021ironman} exploits GNNs to generate graph embeddings of IR graphs, which serve as state representations in its reinforcement learning (RL-)based search engine to find the Pareto curve between two types of computing resources on FPGAs.
The same problem can also be solved by carefully designing a GNN surrogate model as a continuous relaxation of the actual cost model, allowing for a soft solution that can be decoded into the final discrete solution of resource assignments~\cite{wang2022unsupervised}.
In terms of HLS datasets, Wu et al.~\cite{wu2022high} develop an HLS dataset and benchmark GNNs for predicting resource usage and timing, however, they enhance accuracy with domain-specific information and do not explore message passing directions or the benefit from positional encoding. 
Bai et al.~\cite{bai2023towards} combine pre-trained language models~\cite{wang2021codet5, guo2020graphcodebert} and GNNs to predict the optimization effects of different directives.

In logic synthesis or logic design, LOSTIN~\cite{wu2022lostin} employs a GNN to encode circuit graphs and an LSTM to encode logic synthesis sequences, where the two embeddings are concatenated to predict logic delay and area.
To identify functional units from gate-level netlists, different GNN models can be leveraged to classify sub-circuit functionality~\cite{alrahis2021gnn}, predict the functionality of approximate circuits~\cite{bucher2022appgnn}, analyze impacts of circuit rewriting on functional operator detection~\cite{zhao2022graph}, and predict boundaries of arithmetic blocks~\cite{he2021graph}.
Gamora~\cite{wu2023gamora} leverages the message-passing mechanism in GNN computation to imitate structural shape hashing and functional propagation in conventional symbolic reasoning, achieving up to six orders of magnitude speedup compared to the logic synthesis tool ABC in extracting adder trees from multipliers.

In physical synthesis, Mirhoseini et al.~\cite{mirhoseini2021graph} combine GCN with deep RL to place macros (i.e., memory cells), after which standard cells are placed by a force-directed method.
The GCN model encodes the topological information of chip netlists to generate graph embeddings as the inputs to the RL agent, as well as to provide proxy rewards to guide the search process.
Lu et al.~\cite{lu2020vlsi} apply GraphSAGE~\cite{hamilton2017inductive} to circuit netlists to learn node representations that capture logical affinity.
These representations are grouped by a weighted K-means clustering to provide placement guidance, informing the placer about which cells should be placed nearby in actual physical layouts.
Guo et al.~\cite{guo2022timing} develop a hierarchical GNN with BI message passing to estimate post-routing timing behaviors by using circuit placement results.

In hardware design verification, test point insertion is a common technique aimed at enhancing fault coverage, which modifies target hardware designs by inserting extra control points or observation points. Ma et al.~\cite{ma2019high} use GCNs to predict whether a node in hardware designs is easy or hard to observe, based on which new observation points are inserted.
To improve branch coverage, Vasudevan et al.~\cite{vasudevan2021learning} exploit IPA-GNN~\cite{bieber2020learning} to predict the probability of current test parameters covering specific cover points by characterizing RTL semantics and computation flows; new tests targeting uncovered points are generated by maximizing the predicted probability with respect to test parameters through gradient-based search. 

In analog circuit design, by using circuit schemetics, CktGNN~\cite{dong2023cktgnn} employs a nested GNN to predict analog circuit properties (i.e., gain, BW, PM) and reconstruct circuit topology.
By using pre-layout information, ParaGraph~\cite{ren2020paragraph} builds a GNN model to predict layout-dependent parasitics and physical device parameters; GCN-RL circuit designer~\cite{wang2020gcn} combines RL with GCNs for automatic transistor sizing.
By using layout information, GNN surrogate models can predict the relative placement quality of different designs~\cite{li2020customized}, and other circuit properties, such as the electromagnetic properties of high-frequency circuits \cite{zhang2019circuit}.

\section{A Brief Review of Magnetic Laplacian and Positional Encodings for Directed Graphs} \label{app:pe}

Positional encodings (PE) for graphs are vectorized representations that can effectively describe the global position of nodes (absolute PE) or relative position of node pairs (relative PE). They provide crucial positional information and thus benefits many backbone models that is position-agnostic. For instance, on undirected graphs, PE can provably alleviate the limited expressive power of Message Passing Neural Networks~\cite{xu2018powerful, morris2019weisfeiler, li2020distance, lim2022sign}; PE are also widely adopted in many graph transformers to incorporate positional information and break the identicalness of nodes in attention mechanism~\cite{kreuzer2021rethinking, ying2021transformers, rampavsek2022recipe, chen2022structure}. As a result, the design and use of PE become one of the most important factors in building powerful graph encoders.

Likely, one can expect that direction-aware PE are also crucial when it comes to directed graph encoders. ``Direction-aware'' implies that PE should be able to capture the directedness of graphs. A notable example is Magnetic Laplacian PE~\cite{geisler2023transformers}, which adopts the eigenvectors of Magnetic Laplacian as PE. Note that Magnetic Laplacian can encode the directedness via the sign of phase of $\exp\{\pm i2\pi q\}$. Besides, when $q=0$, Magnetic Laplacian reduces to normal symmetric Laplacian. Thus, Magnetic Laplacian PE for directed graphs can be seen as a generalization of Laplacian PE for undirected graphs, and the latter is known to enjoy many nice spectral properties~\cite{chung1997spectral} and be capable to capture many undirected graph distances~\cite{kreuzer2021rethinking}. Therefore, Magnetic Laplacian appears to be a strong candidate for designing direction-aware PE. The definition is as follows:

Magnetic Laplacian (MagLap) matrix is a Hermitian complex matrix defined by $\bm{L}_q = \bm{I} - \bm{D}^{-1/2} \bm{A}_q \bm{D}^{-1/2}$, where $\bm{D}$ is the diagonalized degree matrix counting both in-degree and out-degree, and $\bm{A}_q$ refers to the complex matrix as follows:
\begin{equation}
    [\bm{A}_q]_{u,v} = \left\{ \begin{aligned}
        \exp\{i2\pi q\}&,&\text{if } (u,v)\in \mathcal{E},\\ 
         \exp\{-i2\pi q\}&,&\text{if } (v,u)\in \mathcal{E},\\ 
         1 &, & \text{if } (u,v), (v,u)\in\mathcal{E},
    \end{aligned} \right.
\end{equation}
with a parameter $q\in[0, 1)$ called potential. Hermitian refers to the property that complex conjugate $\bm{L}_q^{\dagger}$ equals to $\bm{L}_q$. It is also worth noticing that when $q=0$, MagLap $\bm{L}_{q=0}$ degenerates to the standard symmetric Laplacian matrix $\bm{L}=\bm{I}-\bm{D}^{-1/2}(\bm{A}+\bm{A}^{\top})\bm{D}^{-1/2}$ as a special case, where $\bm{A}$ is the Adjacency matrix. See~\cite{furutani2020graph} for a comprehensive introduction to Magnetic Laplacian.

Note that it is worth mentioning that there are also other PE for directed graphs, such as SVD of Adjacency matrix~\cite{hussain2022global} and directed random walk~\cite{geisler2023transformers}.

\section{Data split when comparing with baselines in the original papers}
\label{app:data-split}

When comparing with the baselines from original papers, for training and testing the proposed new methods `BI-GINE+EPE' and `BI-GPS+EPE', we follow the dataset split of the original paper for fair comparison.

In the AMP dataset, we follow~\cite{dong2023cktgnn} 
to merge the graphs with 2-stage and 3-stage Op-Amps together into one dataset, we take the last $1000$ graphs for test and the rest for training and validation. The performance of baseline method cktGNN and the proposed new methods `BI-GINE+EPE' and `BI-GPS+EPE' are reported on such data split; for the HLS dataset, both the baseline method and the proposed new methods are trained and tested on control data flow graphs (CDFG) only, following the same data split ratio that randomly divide the data into training, validation and testing as described in the original paper~\cite{wu2022high}; in the SR dataset, both the baseline and the new methods are trained with $24$-bit netlists and tested on $48$-bit netlists, note that both the training and testing data are obtained before technology mapping~\cite{wu2023gamora}; for the CG dataset both the baselines and the proposed methods are tested to predict the runtime of neural networks on the Cortex A76 CPU platform~\cite{zhang2021nn}; for the TIME dataset, we follow the dataset split in the original paper~\cite{guo2022timing} and compare the results of the baseline method and the new methods on the ID designs. 

\section{Dataset Selection Details}
\label{app:dataset_selection_detail}

\textbf{License for the datasets and codes.}

\begin{table}[h]
\centering
\begin{tabular}{@{}ccc@{}}
\toprule
 & code implementation & dataset license \\ \midrule
HLS~\cite{wu2022high} & MIT License & MIT License \\
AMP~\cite{dong2023cktgnn} & MIT License & MIT License \\
SR~\cite{wu2023gamora} & The MIT License & The MIT License \\
CG~\cite{zhang2021nn} & MIT License & MIT License \\
our benchmark & CC BY-NC & - - \\ \bottomrule
\end{tabular}
\caption{License of the datasets and the toolbox implementation of this benchmark.}
\label{tab:license}
\end{table}

For detailed information of the license of each origin dataset, please refer to their original paper/documents, the final interpretation regarding the five dataset's licensing information rests with the owner of the original paper. To the best of our knowledge, these hardware datasets contain no personally identifiable information or offensive content.

\textbf{Detailed Description of Evaluation Metrics}

\subsection{High-Level Synthesis (HLS) Dataset}
\label{app:hls_description}

\begin{figure}[h]
  \centering
  \begin{minipage}[t]{0.36\textwidth}
    \centering
    \includegraphics[width=\linewidth]{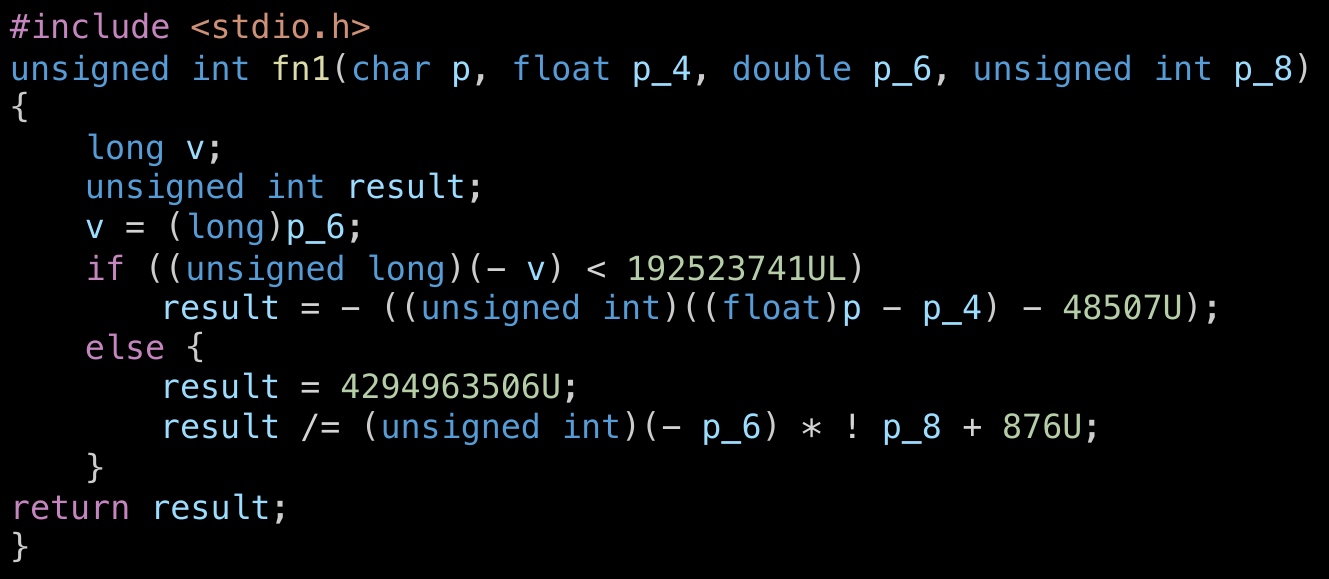}
    \caption{An example C program from the CDFG dataset.}
    \label{fig:hls_cdfg:a}
  \end{minipage}
  \hfill
  \begin{minipage}[t]{0.53\textwidth}
    \centering
    \includegraphics[width=\linewidth]{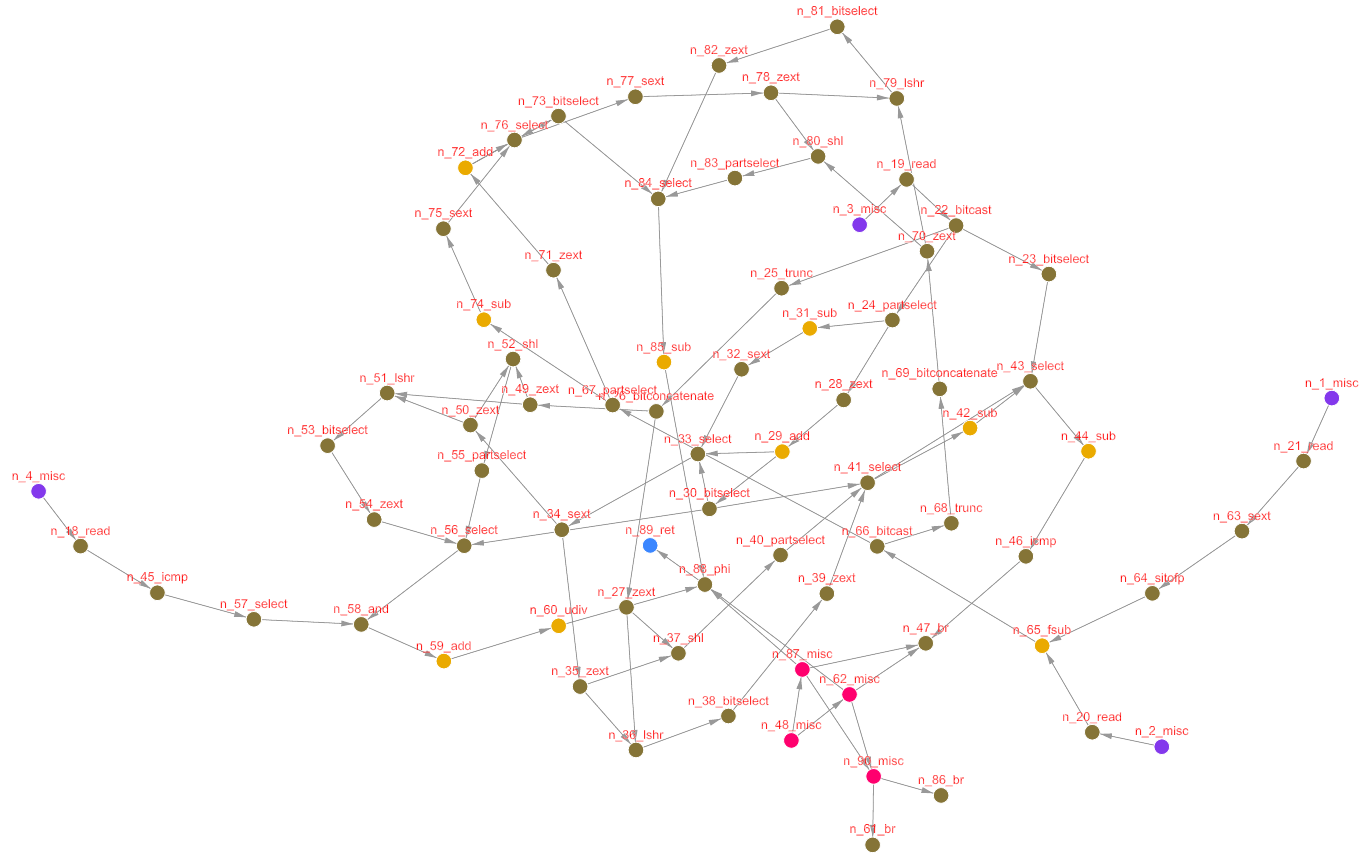}
    \caption{Control dataflow graph of the example program in Figure~\ref{fig:hls_cdfg:a}.}
    \label{fig:hls_cdfg:b}
  \end{minipage}
\end{figure}

After HLS front-end compilation, six node features are extracted, as summarized in Table~\ref{table:appendix_hls}.
Each edge has two features, the edge type represented in integers, and a binary value indicating whether this edge is a back edge.
Each graph is labeled based on its post-implementation performance metrics, which are synthesized by Vitis HLS~\cite{vitishls} and implemented by Vivado \cite{vivado}.
Three metrics are used for regression: \texttt{DSP}, \texttt{LUT}, and \texttt{CP}.
The first two are integer numbers indicating the number of resources used in the final implementation; the last one is \texttt{CP} timing in fractional number, determining the maximum working frequency of FPGA.
The DFG and CDFG datasets consists of 19,120 and 18,570 C programs, respectively.
Figure~\ref{fig:hls_cdfg:a} shows an example C program from the CDFG dataset, with the corresponding control dataflow graph shown in Figure~\ref{fig:hls_cdfg:b}.
More information can be found in the original paper~\cite{wu2022high}.

\begin{center}
\resizebox{\textwidth}{!}{\begin{tabular}{@{}ccc@{}}
\toprule
\textbf{Feature} & \textbf{Description} & \textbf{Values} \\ \midrule
Node type   &  General node type & \texttt{operation nodes}, \texttt{blocks}, \texttt{ports}, \texttt{misc}\\ 
\rowcolor[HTML]{EFEFEF}
Bitwidth          &  Bitwidth of the node  & \texttt{0}$\sim$\texttt{256}, \texttt{misc}\\
Opcode type   &  Opcode categories based on LLVM  & \texttt{binary\_unary}, \texttt{bitwise}, \texttt{memory}, etc.  \\ 
\rowcolor[HTML]{EFEFEF}
Opcode            &  Opcode of the node &  \texttt{load}, \texttt{add}, \texttt{xor}, \texttt{icmp}, etc. \\
Is start of path  & Whether the node is the starting node of a path  & \texttt{0}, \texttt{1}, \texttt{misc}   \\
\rowcolor[HTML]{EFEFEF}

Cluster group  & Cluster number of the node & \texttt{-1}$\sim$\texttt{256}, \texttt{misc}\\
\bottomrule
\end{tabular}}
\captionof{table}{Node features and their example values.}
\label{table:appendix_hls}
\end{center}

\subsection{Symbolic Reasoning (SR) Dataset}
\label{app:sr_description}

In this dataset, all the circuit designs are represented as and-inverter graphs (AIGs), a concise and uniform representation of BNs consisting of inverters and two-input AND gates, which allows rewriting, simulation, technology mapping, placement, and verification to share the same data structure~\cite{mishchenko2006dag}.
In an AIG, each node has at most two incoming edges; 
a node without incoming edges is a primary input (PI);
primary outputs (POs) are denoted by special output nodes;
each internal node represents a two-input AND function. 
Based on De Morgan’s laws, any combinational BN can be converted into an AIG~\cite{brayton2010abc} in a fast and scalable manner.

For each node, there are three node features represented in binary values denoting node types and Boolean functionality.
The first node feature indicates whether this node is a PI/PO or intermediate node (i.e., AND gate).
The second and the third node features indicate whether each input edge is inverted or not, such that AIGs can be represented as homogeneous graphs without additional edge features.

\begin{figure}[h]
    \centering
    \includegraphics[width=\textwidth]{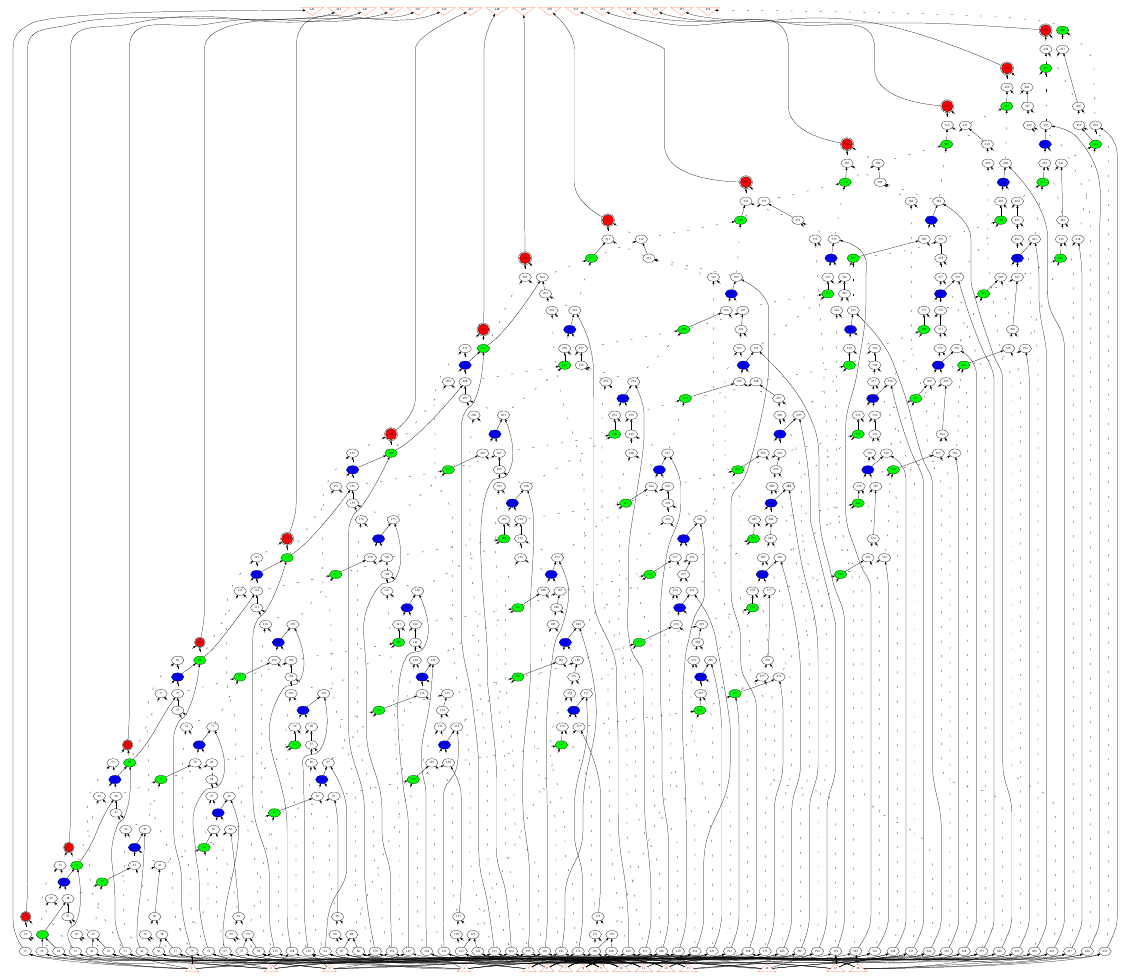}
    \caption{8-bit multiplier in AIG.}
    \label{fig:sr_mult8_aig}
\end{figure}

\begin{figure}[h]
    \centering
    \includegraphics[width=\textwidth]{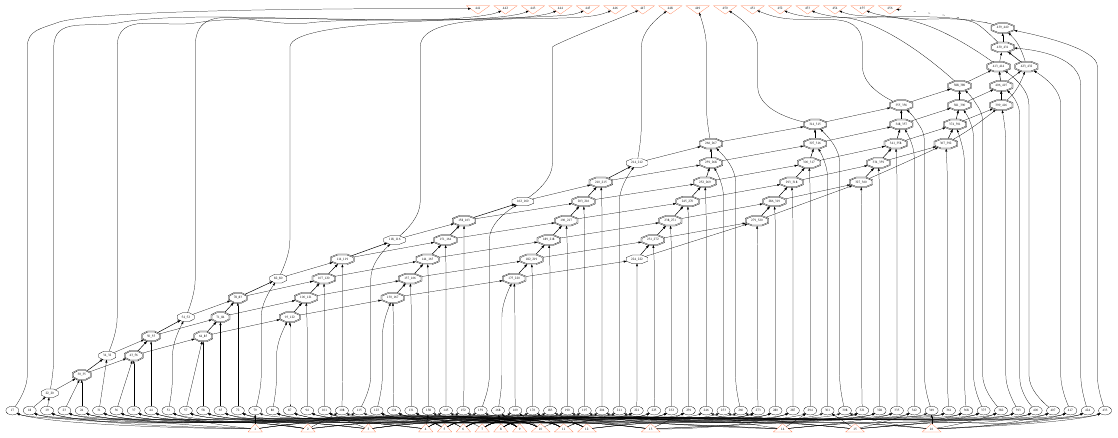}
    \caption{8-bit multiplier with adders extracted.}
    \label{fig:sr_mult8_adder}
\end{figure}

This dataset aims to leverage graph learning based approaches to accelerate the adder tree extraction in (integer) multiplier verification, which involves two reasoning steps~\cite{li2013wordrev,subramanyan2013reverse}:
(1) detecting XOR/MAJ functions to construct adders, and then (2) identifying their boundaries. 
Thus, there are two sets of node labels, i.e., two node-level classification tasks.
One task provides labels specifying whether a node (i.e., a gate) in the AIG belongs to MAJ, XOR, or is shared by both MAJ and XOR.
The other task provides labels specifying whether a node is the root node of an adder.
These AIGs and ground truth labels are generated by the logic synthesis tool ABC~\cite{brayton2010abc}.
Figure~\ref{fig:sr_mult8_aig} shows the AIG of an 8-bit multiplier: the blue and red nodes are the root nodes of XOR functions, with the red nodes directly connecting to the POs; the green nodes are the root nodes of MAJ functions.
By pairing one XOR function with one MAJ function sharing the same set of inputs, we can extract the adder tree, which is shown in Figure~\ref{fig:sr_mult8_adder}.
More information can be found in the original paper~\cite{wu2023gamora}.

\subsection{Pre-routing Timing Prediction (TIME) Dataset}
\label{app:time_description}

Similar to timing analysis tools, circuits in this dataset are represented as heterogeneous graphs consisting of two types of edges: net edges and cell edges, with edge features shown in Table~\ref{table:appendix_time_edgefeat}.
The nodes in graphs denote pins in circuits, with features summarized in Table~\ref{table:appendix_time_nodefeat}.
The TIME dataset collects 21 real-world benchmark circuits from OpenCores~\cite{opencores} with OpenROAD~\cite{openroad} on SkyWater $130$nm technology~\cite{skywater} (i.e. blabla, usb\_cdc\_core, BM64, salsa20, aes128, aes192, aes256, wbqspiflash, cic\_decimator, des, aes\_cipher, picorv32a, zipdiv, genericfir, usb, jpeg\_encoder, usbf\_device, xtea, spm, y\_huff, and synth\_ram).
More information can be found in the original paper~\cite{guo2022timing}.


\begin{table}
\begin{subtable}{.5\linewidth}
        \centering
        \resizebox{\textwidth}{!}{
        \begin{tabular}{cc}
        \toprule
        \textbf{Description} & \textbf{Size} \\ \midrule
        (Net edge) Distances along x/y direction   &  2\\ 
        \rowcolor[HTML]{EFEFEF}
        (Cell edge) LUT is valid or no      &  8\\
        (Cell) LUT indices           &  $8\times(7+7)$ \\ 
        \rowcolor[HTML]{EFEFEF}
        (Cell) LUT value matrices    &  $8\times(7\times7)$ \\
        \bottomrule
        \end{tabular}}
        \caption{Edge features in the TIME dataset. For each cell edge, 8 LUTs are used to model cell delay and slew under four timing corner combinations (EL/RF).}
        \label{table:appendix_time_edgefeat}
\end{subtable} 
\hfill
\begin{subtable}{.46\linewidth}
        \centering
        \resizebox{\textwidth}{!}{
        \begin{tabular}{cc}
        \toprule
        \textbf{Description} & \textbf{Size} \\ \midrule
        Is primary I/O pin or not   &  1\\ 
        \rowcolor[HTML]{EFEFEF}
        Is fan-in or fan-out          &  1\\
        Distance to the 4 die area boundaries   &  4  \\ 
        \rowcolor[HTML]{EFEFEF}
        Pin capacitance            &  4 (EL/RF) \\
        \bottomrule
        \end{tabular}}
        \caption{Pin (i.e., node) features in the TIME dataset. EL/RF stands for early/late and rise/fall, i.e., the four timing corner combinations in STA.}
        \label{table:appendix_time_nodefeat}
\end{subtable}%
\caption{Node and edge features for pre-routing timing prediction.}
\end{table}

We select the slack prediction task in this dataset, including setup slack and hold slack.
Slack values are used by STA tools to identify paths that violate timing constraints, enabling further optimization of placement and routing.
Setup/hold slack is defined as the difference between the required arrival time (based on setup or hold time) and the actual arrival time of data/signals at timing endpoints, making it a node-level regression task.

\begin{figure}[h]
    \centering
    \includegraphics[width=0.5\textwidth]{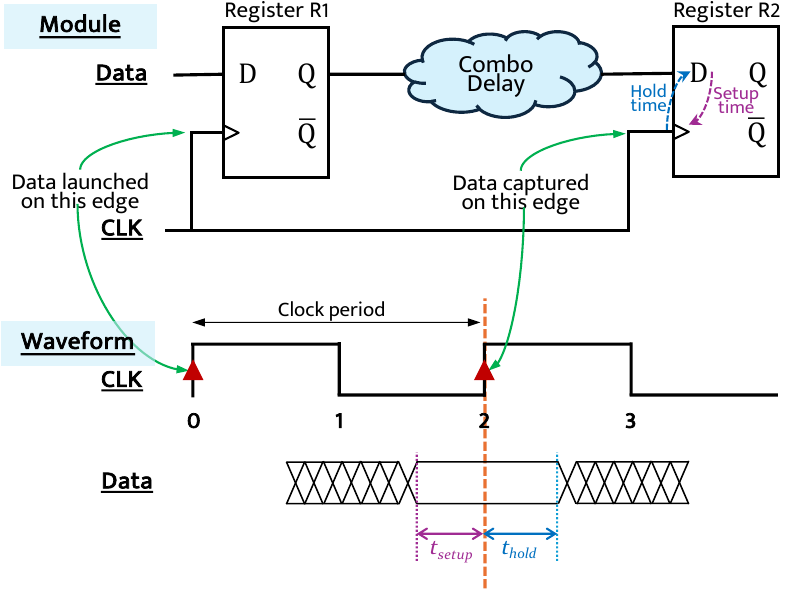}
    \caption{Register-to-register timing path.}
    \label{fig:time_slack}
\end{figure}

Figure~\ref{fig:time_slack} shows the most common timing path, register-to-register path.
(1) For setup slack, the signal should arrive \textit{earlier} than the required arrival time (i.e., clock period - setup time).
Setup time $t_{\text{setup}}$ refers to the time before the clock edge that data must be stable.
(2) For hold slack, the signal should arrive \textit{later} than the required hold time to ensure no impact on signals for the current clock edge. 
Hold time $t_{\text{hold}}$ refers to the time after the clock edge that data must be stable.

\subsubsection{TIME Dataset Distribution Shift Definition} 
\label{app:time_distirbution_definition}

For training and ID testing, we take the designs `blabla', `usb\_cdc\_core', `wbqspiflash', `cic\_decimator', `picorv32a', `zipdiv', `usb'. 
For OOD testing, we use `xtea', `spm', `y\_huff', `synth\_ram'.

\subsection{Computational Graph (CG) Dataset}
\label{app:cg_description}

\begin{figure}[h]
    \centering
    \includegraphics[width=0.65\textwidth]{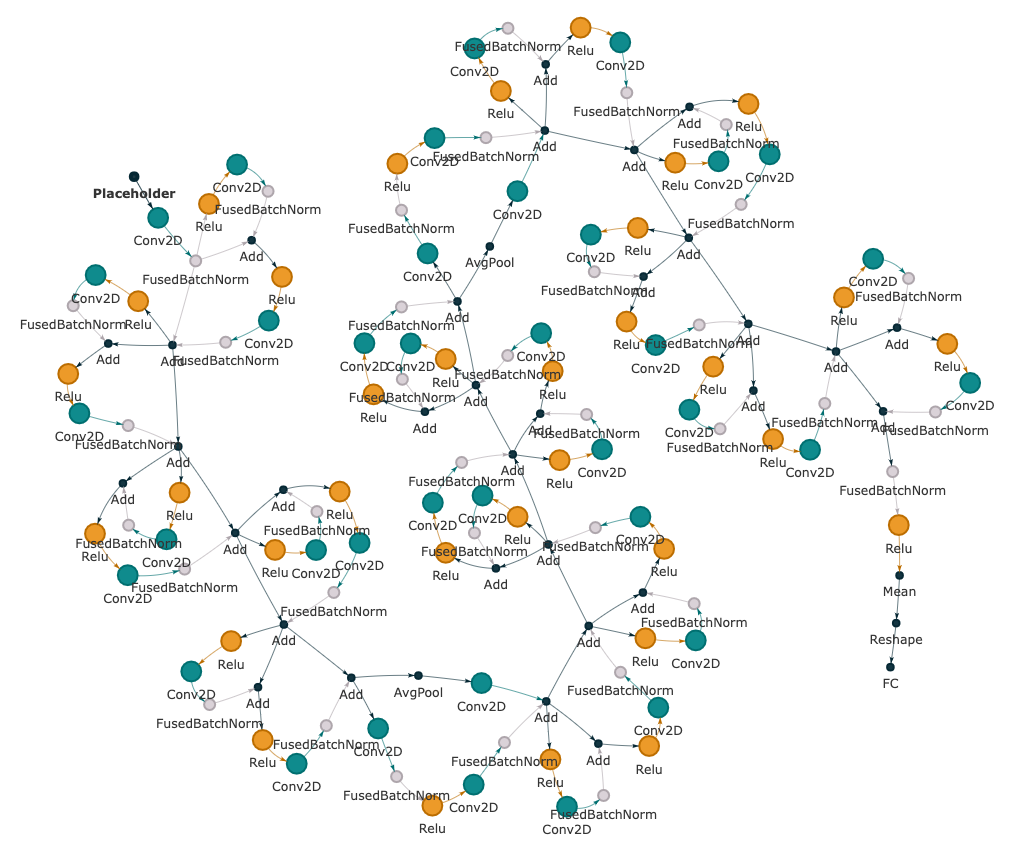}
    \caption{Computational graph of an example NN model from NASBench201~\cite{dong2019bench}.}
    \label{fig:cg_example}
\end{figure}

This dataset includes (1) 12 state-of-the-art CNN models for the ImageNet2012 classification task (i.e., AlexNet, VGG, DenseNet, ResNet, SqueezeNet, GoogleNet, MobileNetv1,  MobileNetv2,  MobileNetv3, ShuffleNetv2, MnasNet, and ProxylessNas), each with 2,000 variants that differ in output channel number and kernel size per layer, and (2) 2,000 models from NASBench201~\cite{dong2019bench} with the highest test accuracy on CIFAR10, each featuring a unique set of edge connections.
In total, this dataset contains 26,000 models with different operators and configurations.
Figure~\ref{fig:cg_example} shows an example of the computational graph of a model in NASBench201.

Node features include input shape (5 dimensions), kernel/weight shape (padding to 4 dimensions), strides (2 dimensions), and output shape (5 dimensions).
Each computational graph is labeled with the inference latency on three edge devices (i.e., Cortex A76 CPU, Adreno 630 GPU, Adreno 640 GPU).
There is no edge feature in this dataset.
More information can be found in the original paper~\cite{zhang2021nn}.

\subsubsection{CG Dataset Distribution Shift Definition}
\label{app:cg_distribution_definition}

For training and ID testing, we take `DenseNets', `MnasNets', `MobileNetv2s', `MobileNetv3s', `nasbench201s'.
For OOD testing, we select `Proxylessass', `ResNets', and `SqueezeNets'.

\subsection{Multi-Stage Amplifiers (AMP) Dataset}
\label{app:amp_description}

\begin{figure}[h]
    \centering
    \includegraphics[width=\textwidth]{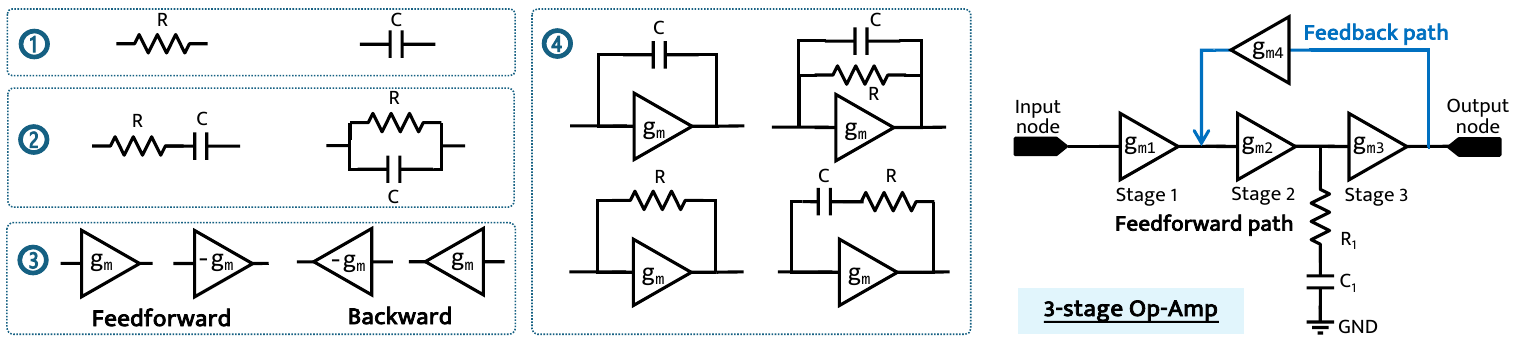}
    \caption{Subgraph basis for operational amplifiers and an example 3-stage Op-Amp.}
    \label{fig:analog_example}
\end{figure}

This dataset focuses on predicting circuit specifications (e.g., DC gain, bandwidth (BW), phase margin (PM)) of 2/3-stage operational amplifiers (Op-Amps), which are simulated by the circuit simulator Cadence Spectre~\cite{spectre}.
A 2/3-stage Op-Amp consists of (1) two/three single-stage Op-Amps on the main feedforwoard path and (2) several feedback paths, with one example shown in the right part of Figure~\ref{fig:analog_example}.
To make multi-stage Op-Amps more stable, feedforward and feedback paths are used to achieve different compensation schemes, each of which is implemented with a sub-circuit, e.g., single-stage Op-Amps, resistors, and capacitors.
Due to the different topologies of single-stage Op-Amps and various compensation schemes, each sub-circuit is built as a subgraph.
There are 24 potential sub-circuits in the considered 2/3-stage Op-Amps:
\begin{itemize}
    \item Single R or C (\circled{1} in Figure~\ref{app:amp_description}, 2 types).
    \item R and C connected in parallel or serial (\circled{2} in Figure~\ref{app:amp_description}, 2 types).
    \item A single-stage Op-Amp ($g_m$) with different polarities (positive, $+g_m$, or negative, $-g_m$) and directions (feedforward or feedback) (\circled{3} in Figure~\ref{app:amp_description}, 4 types).
    \item A single-stage Op-Amp ($g_m$) with R or C connected in parallel or serial (16 types). Note that we use the single-stage Op-Amp with feedforward direction and positive polarities as an example for \circled{4} in Figure~\ref{app:amp_description}.
\end{itemize}

Based on aforementioned formulation, node features include (1) subgraph type, (2) node type (e.g., R, C, $\pm g_m$ with feedforward/feedback, primary input/output), and (3) value of the component.
There is no edge feature.
More information can be found in the original paper~\cite{dong2023cktgnn}.

\section{Benchmark Design Details}

\subsection{Selected Backbone Functional}
\label{app:gnn_backbone}
Here we list the functions we implemented for the selected GNN backbone layers, note that here we show the forms of the backbone on undirected graphs, one may do slight modification by introducing $\omega(\cdot)$ on the neighbor message aggregation to consider message passing control for directed graphs.
\begin{equation}
    \text{GIN}: \ \ \mathbf{x}_i^{(k)} = \text{MLP} \left(\mathbf{x}_i^{(k-1)} + \sum_{j \in \mathcal{N}(i)} \mathbf{x}_j^{(k-1)} \right)
\end{equation} is the for graphs without edge features, 

\begin{equation}
    \text{GINE}: \ \ \mathbf{x}^{(k)}_i = \text{MLP} \left(
\mathbf{x}_i^{(k-1)} + \sum_{j \in \mathcal{N}(i)} \mathrm{ReLU}
( \mathbf{x}_j^{(k-1)} + \mathbf{e}_{j,i}^{(k-1)} ) \right)
\end{equation} is used for graphs with edge features.

\begin{equation}
    \text{GCN:} \ \ \mathbf{x}^{(k)}_i = \theta^{\top} \sum_{j \in
\mathcal{N}(i) \cup \{ i \}} \frac{\mathbf{e}_{j,i}^{(k-1)}}{\sqrt{\hat{d}_j
\hat{d}_i}} \mathbf{x}_j^{(k-1)},
\end{equation}where $\theta$ is the parameter to learn, for graphs with edge features $\mathbf{e}_{j,i}$ is the processed edge weight, for graphs without edge features $\mathbf{e}_{j,i}$ is set as $1$.

\begin{equation}
    \text{GAT:} \ \ \mathbf{x}^{(k)}_i = \alpha_{i,i}^{(k-1)} \theta_{s}\mathbf{x}_{i}^{(k-1)} +
\sum_{j \in \mathcal{N}(i)}
\alpha_{i,j}^{(k-1)} \theta_{t}\mathbf{x}_{j}^{(k-1)},
\end{equation}where $\theta_s, \theta_t$ are parameters to learn, for graphs without edge features, 
\begin{equation}
    \alpha_{i,j}^{(k-1)} =
\frac{
\exp\left(\mathrm{LeakyReLU}\left(
\mathbf{a}^{\top}_{s} \theta_{s}\mathbf{x}_i^{(k-1)}
+ \mathbf{a}^{\top}_{t} \theta_{t}\mathbf{x}_j^{(k-1)}
\right)\right)}
{\sum_{m \in \mathcal{N}(i) \cup \{ i \}}
\exp\left(\mathrm{LeakyReLU}\left(
\mathbf{a}^{\top}_{s} \theta_{s}\mathbf{x}_i^{(k-1)}
+ \mathbf{a}^{\top}_{t}\theta_{t}\mathbf{x}_m^{(k-1)}
\right)\right)},
\end{equation}and for graphs with edge features,
\begin{equation}
    \alpha_{i,j}^{(k-1)} =
\frac{
\exp\left(\mathrm{LeakyReLU}\left(
\mathbf{a}^{\top}_{s} \theta_{s}\mathbf{x}_i^{(k-1)}
+ \mathbf{a}^{\top}_{t} \theta_{t}\mathbf{x}_j^{(k-1)}
+ \mathbf{a}^{\top}_{e} \theta_{e} \mathbf{e}_{i,j}^{(k-1)}
\right)\right)}
{\sum_{m \in \mathcal{N}(i) \cup \{ i \}}
\exp\left(\mathrm{LeakyReLU}\left(
\mathbf{a}^{\top}_{s} \theta_{s}\mathbf{x}_i^{(k-1)}
+ \mathbf{a}^{\top}_{t} \theta_{t}\mathbf{x}_m^{(k-1)}
+ \mathbf{a}^{\top}_{e} \theta_{e} \mathbf{e}_{i,m}^{(k-1)}
\right)\right)},
\end{equation}where $\mathbf{a}_s, \mathbf{a}_t$ are learnable parameterized attention parameters.

Each GPS backbone layer is implemented as follows:
\begin{equation}
\begin{aligned}
    \text{GPS}: & \mathbf{X}^{(k)}_M = \text{MPNN}^{(k-1)}(\mathbf{X}^{(k-1)}, \mathbf{E}^{(k-1)}) \\ & \mathbf{X}^{(k)}_T = \text{GlobalATTn}^{(k-1)}(\mathbf{X}^{(k-1)} \\
    & \mathbf{X}^{(k)} = \text{MLP}(\mathbf{X}^{(k-1)}_M + \mathbf{X}^{(k-1)}_T),
\end{aligned}
\end{equation}
where $\mathbf{X}, \mathbf{E}$ denote node/edge features, we use GIN or GINE as the MPNN layer, and we use the transformer as the global attention reasoning layer.

For DGCN~\cite{tong2020directed} and DiGCN~\cite{tong2020digraph}, we follow the implementation in PyGSD~\cite{he2024pytorch}, please refer to \url{https://pytorch-geometric-signed-directed.readthedocs.io/en/latest/index.html} for backbone implementation details.

\begin{equation}
    \text{MSGNN:} \ \ \mathbf{x}_i^{(k)} = \sigma \left( \sum_{i=1}^{F^{k-1}} \mathbf{Y}_{ij} \mathbf{x}_i^{(k-1)} + \mathbf{b}_j^{(k-1)} \right),
\end{equation}where $\sigma$ is a complex version of Rectified Linear Unit defined by:
\[
\sigma(z) = 
\begin{cases} 
z & -\pi / 2 \leq \arg (z) < \pi / 2 \\
0 & \text{otherwise},
\end{cases}
\]
where $\arg(\cdot)$ is the complex argument of $z \in \mathbb{C}$, $F^{(k)}$ denotes the number of channels in the $k$-th layer, $\mathbf{b}$ is a bias vector with equal real and imaginary parts, $\mathbf{Y}$ denotes the convolution matrix defined in Equation.(4) and (5) in ~\cite{he2022msgnn}. 

\subsection{Hyper-Parameter Space}
\label{app:hyper-parameter_space}

\begin{table}[h]

\resizebox{\textwidth}{!}{\begin{tabular}{@{}ccccccc@{}}
\toprule
 & batch size & learning rate & dropout rate & hidden dimension$^*$ & \# of GNN layers & \# of MLP layers \\ \midrule
DGCN & \{64, 128, 256, 512, 1024\} & [5e-4, 1e-2] & \{0, 0.1, 0.2, 0.3\} & [96, 336] & [3,8] & [2,5] \\
DiGCN & \{64, 128, 256, 512, 1024\} & [5e-4, 1e-2] & \{0, 0.1, 0.2, 0.3\} & [96, 336] & [3,8] & [2,5] \\
MagNet & \{64, 128, 256, 512, 1024\} & [5e-4, 1e-2] & \{0, 0.1, 0.2, 0.3\} & [96, 336] & [3,8] & [2,5] \\
GCN & \{64, 128, 256, 512, 1024\} & [1e-4, 1e-2] & \{0, 0.1, 0.2, 0.3\} & [96, 336] & [3,8] & [2,5] \\ \midrule
GIN & \{64, 128, 256, 512, 1024\} & [1e-4, 1e-2] & \{0, 0.1, 0.2, 0.3\} & [96, 336] & [3,8] & [2,5] \\
GAT & \{64, 128, 256, 512, 1024\} & [1e-4, 1e-2] & \{0, 0.1, 0.2, 0.3\} & [96, 336] & [3,8] & [2,5] \\ \midrule
GPS-T & \{64, 128, 256\} & [1e-4, 1e-2] & \{0, 0.1, 0.2, 0.3\} & [96, 288] & [3,6] & [2,5] \\
GPS-P & \{32, 64, 128, 256, 512\} & [1e-4, 1e-2] & \{0, 0.1, 0.2, 0.3\} & [96, 288] & [3,6] & [2,5] \\ \bottomrule
\end{tabular}}
\caption{Hyper-parameter space for each backbone. $^*$:hidden dimension slightly vary in each task.}
\label{tab:hyper-parameter_space}
\end{table}

\section{Hardware and Platform}
All the experiments run on a server with an AMP EPYC 7763 64-Core Processor and 8 Nvidia RTX6000 GPU cards. The codes run on frameworks based on PyTorch~\cite{paszke2019pytorch}, PyTorch Geometric~\cite{Fey/Lenssen/2019}, PyTorch Geometrc Signed and Directed~\cite{he2024pytorch}, RAY~\cite{liaw2018tune}.

\section{Implementation Details of Experiments}
\subsection{Ranking Calculation}
\label{app:ranking_expression}
In Table.~\ref{tab:exp_gnn_backbone_selection} and Table.~\ref{tab:node_edge_PE}, we report the average ranking of different combination of methods w.r.t. per evaluation metrics for each task from each dataset. The calculation of the ranking can be expressed as:
\begin{equation}
\label{equ:gnn_backbone_ranking_calculation_expression}
\begin{aligned}
    \text{rank}_k^{t, D} = \frac{1}{M_{D}} \sum_{m=1}^{M_{D}} R_{t,m}^k,
\end{aligned}
\end{equation}where $R_{t,m}^k$ denotes the ranking of the DGRL method $k$ on task $t$ w.r.t. the $m$-th evaluation metric. $M_{D}$ denotes the number of tasks and metrics on dataset $D$. 

For evaluation metric the larger the better, we adopt the ranking function from pandas~\cite{reback2020pandas} with parameter $ascending = Flase$ and $method = `max'$.

For evaluation metrics the smaller the better, we use $ascending = True$ and $method = `min'$.

\clearpage
\section{Detailed Experiment Results}
\subsection{Main Results: In-distribution and Out-of-distribution Performance}
\label{app:exp_main_result}
\begin{center}
\resizebox{0.9\textwidth}{!}{
\begin{tabular}{@{}ccccccccc@{}}
\toprule
\multirow{2}{*}{Method} & \multicolumn{4}{c}{shared} & \multicolumn{4}{c}{root} \\ \cmidrule(l){2-9} 
 & accuracy & precision & recall & f1 & accuracy & precision & recall & f1 \\ \midrule
GCN & 0.879±0.013 & 0.669±0.141 & 0.653±0.125 & 0.620±0.119 & 0.882±0.005 & 0.860±0.045 & 0.811±0.131 & 0.773±0.091 \\
DI-GCN & 0.633±0.000 & 0.376±0.000 & 0.377±0.000 & 0.337±0.000 & 0.751±0.000 & 0.250±0.000 & 0.333±0.000 & 0.285±0.000 \\
BI-GCN & 0.992±0.005 & 0.816±0.123 & 0.820±0.122 & 0.818±0.123 & 0.999±0.001 & 0.999±0.000 & 0.998±0.003 & 0.999±0.001 \\
GIN & 0.882±0.046 & 0.787±0.132 & 0.704±0.191 & 0.683±0.179 & 0.909±0.005 & 0.900±0.055 & 0.877±0.101 & 0.850±0.050 \\
DI-GIN & 0.999±0.000 & 0.749±0.000 & 0.749±0.000 & 0.749±0.000 & 0.999±0.000 & 0.999±0.000 & 0.999±0.000 & 0.999±0.000 \\
BI-GIN & 0.999±0.000 & 0.937±0.088 & 0.974±0.079 & 0.949±0.080 & 0.999±0.000 & 0.999±0.000 & 0.999±0.000 & 0.999±0.000 \\
\bottomrule
\end{tabular}}
\captionof{table}{ID performance on the SR dataset.}
\label{tab:sr_id}
\end{center}


\begin{center}
\resizebox{0.9\textwidth}{!}{\begin{tabular}{@{}ccccccc@{}}
\toprule
\multirow{2}{*}{Method} & \multicolumn{2}{c}{gain} & \multicolumn{2}{c}{PM} & \multicolumn{2}{c}{BW} \\ \cmidrule(l){2-7} 
 & mse & rmse & mse & rmse & mse & rmse \\ \midrule
GCN & 1.262±1.682 & 0.993±0.554 & 13.598±30.906 & 2.618±2.736 & 35.230±0.657 & 5.935±0.055 \\
DI-GCN & 0.337±0.002 & 0.580±0.002 & 1.243±0.014 & 1.115±0.006 & 36.302±0.245 & 6.025±0.020 \\
BI-GCN & 0.148±0.004 & 0.385±0.005 & 63.874±194.025 & 3.862±7.375 & 22.947±1.022 & 4.789±0.105 \\
GIN & 0.166±0.029 & 0.406±0.034 & 1.266±0.034 & 1.125±0.015 & 28.259±10.648 & 5.244±0.894 \\
DI-GIN & 0.200±0.136 & 0.433±0.117 & 1.300±0.024 & 1.140±0.010 & 24.091±1.270 & 4.906±0.130 \\
BI-GIN & 0.137±0.012 & 0.370±0.016 & 1.251±0.035 & 1.118±0.016 & 19.724±1.489 & 4.438±0.170 \\
GAT & 0.158±0.008 & 0.397±0.010 & 865.339±2684.901 & 11.663±28.466 & 22.770±1.045 & 4.770±0.111 \\
DI-GAT & 0.205±0.006 & 0.453±0.007 & 1.562±0.840 & 1.223±0.268 & 26.855±2.928 & 5.175±0.284 \\
BI-GAT & 0.138±0.007 & 0.372±0.010 & 1.213±0.055 & 1.101±0.024 & 30.333±13.386 & 5.409±1.088 \\
GPS-T & 0.405±0.022 & 0.636±0.017 & 1.277±0.072 & 1.129±0.031 & 16.758±0.754 & 4.092±0.093 \\
DI-GPS-T & 0.122±0.009 & 0.349±0.013 & 1.259±0.044 & 1.121±0.019 & 16.600±0.877 & 4.073±0.107 \\
BI-GPS-T & 0.122±0.007 & 0.349±0.010 & 1.212±0.058 & 1.100±0.026 & 20.475±8.853 & 4.456±0.825 \\ \midrule
DGCN & 0.567±0.004 & 0.753±0.003 & 1.292±0.000 & 1.136±0.000 & 54.256±0.257 & 7.365±0.017 \\
DiGCN & 0.367±0.009 & 0.606±0.007 & 1.294±0.011 & 1.137±0.005 & 52.375±0.276 & 7.237±0.019 \\
MagNet & 0.185±0.008 & 0.431±0.009 & 1.315±0.082 & 1.146±0.035 & 24.800±2.834 & 4.972±0.283 \\ \bottomrule
\end{tabular}}
\captionof{table}{ID performance on the AMP dataset.}
\label{tab:amp_id}
\end{center}
\begin{center}
\resizebox{0.9\textwidth}{!}{\begin{tabular}{@{}ccccccc@{}}
\toprule
\multirow{2}{*}{Method} & \multicolumn{2}{c}{DSP} & \multicolumn{2}{c}{LUT} & \multicolumn{2}{c}{CP} \\ \cmidrule(l){2-7} 
 & mse & R2 & mse & R2 & mse & R2 \\ \midrule
GCN & 12.700±0.324 & 0.877±0.004 & 4.909±0.123 & 0.647±0.021 & 0.713±0.037 & 0.829±0.012 \\
DI-GCN & 12.591±0.312 & 0.877±0.003 & 4.998±0.114 & 0.643±0.013 & 0.692±0.013 & 0.837±0.004 \\
BI-GCN & 10.285±0.336 & 0.902±0.004 & 4.311±0.149 & 0.732±0.010 & 0.665±0.025 & 0.847±0.007 \\
GINE & 2.707±0.133 & 0.975±0.001 & 2.172±0.108 & 0.861±0.008 & 0.653±0.014 & 0.849±0.003 \\
DI-GINE & 2.312±0.172 & 0.979±0.001 & 2.145±0.158 & 0.863±0.011 & 0.645±0.022 & 0.851±0.007 \\
BI-GINE & 2.137±0.076 & 0.981±0.000 & 1.759±0.087 & 0.892±0.005 & 0.629±0.020 & 0.855±0.005 \\
GAT & 4.680±0.264 & 0.957±0.002 & 3.267±0.142 & 0.778±0.011 & 0.643±0.012 & 0.850±0.004 \\
DI-GAT & 7.697±0.238 & 0.926±0.002 & 4.188±0.226 & 0.685±0.031 & 0.677±0.051 & 0.840±0.014 \\
BI-GAT & 4.718±0.532 & 0.957±0.004 & 3.028±0.143 & 0.801±0.016 & 0.590±0.011 & 0.863±0.006 \\
GPS & 2.444±0.207 & 0.978±0.002 & 2.114±0.153 & 0.872±0.011 & 0.621±0.028 & 0.858±0.010 \\
DI-GPS & 2.517±0.180 & 0.977±0.001 & 2.306±0.224 & 0.862±0.015 & 0.625±0.028 & 0.856±0.007 \\
BI-GPS & 2.442±0.303 & 0.979±0.002 & 2.112±0.216 & 0.873±0.014 & 0.621±0.018 & 0.859±0.009 \\ \midrule
DGCN & 19.614±1.151 & 0.816±0.010 & 7.988±2.512 & 0.333±0.256 & 1.127±0.049 & 0.706±0.014 \\
DiGCN & 12.125±0.204 & 0.885±0.003 & 5.683±0.638 & 0.527±0.092 & 0.704±0.019 & 0.836±0.004 \\
MagNet & 4.375±0.452 & 0.961±0.003 & 2.381±0.175 & 0.848±0.015 & 0.684±0.045 & 0.843±0.014 \\ \bottomrule
\end{tabular}}
\captionof{table}{ID performance on the HLS dataset.}
\label{tab:hls_id}
\end{center}

\clearpage

\begin{center}
\resizebox{0.9\textwidth}{!}{\begin{tabular}{@{}ccccccccc@{}}
\toprule
\multirow{2}{*}{Method} & \multicolumn{4}{c}{shared} & \multicolumn{4}{c}{root} \\ \cmidrule(l){2-9} 
 & accuracy & precision & recall & f1 & accuracy & precision & recall & f1 \\ \midrule
GCN & 0.510±0.021 & 0.283±0.014 & 0.328±0.018 & 0.292±0.016 & 0.569±0.056 & 0.364±0.069 & 0.317±0.038 & 0.321±0.038 \\
DI-GCN & 0.553±0.000 & 0.318±0.000 & 0.374±0.000 & 0.301±0.000 & 0.741±0.000 & 0.247±0.000 & 0.333±0.000 & 0.283±0.000 \\
BI-GCN & 0.671±0.045 & 0.489±0.035 & 0.563±0.045 & 0.513±0.037 & 0.651±0.054 & 0.512±0.057 & 0.514±0.064 & 0.505±0.062 \\
GIN & 0.677±0.039 & 0.414±0.034 & 0.395±0.064 & 0.371±0.067 & 0.743±0.008 & 0.646±0.084 & 0.378±0.027 & 0.367±0.045 \\
DI-GIN & 0.621±0.027 & 0.449±0.047 & 0.462±0.024 & 0.441±0.026 & 0.729±0.010 & 0.475±0.047 & 0.390±0.079 & 0.371±0.082 \\
BI-GIN & 0.712±0.024 & 0.514±0.040 & 0.561±0.071 & 0.500±0.049 & 0.773±0.018 & 0.592±0.070 & 0.502±0.051 & 0.515±0.054 \\
\bottomrule
\end{tabular}}
\captionof{table}{OOD performance on the SR dataset.}
\label{tab:sr_ood}
\end{center}
\begin{center}
\resizebox{0.9\textwidth}{!}{\begin{tabular}{@{}ccccccc@{}}
\toprule
\multirow{2}{*}{Method} & \multicolumn{2}{c}{gain} & \multicolumn{2}{c}{pm} & \multicolumn{2}{c}{bw} \\ \cmidrule(l){2-7} 
 & mse & rmse & mse & rmse & mse & rmse \\ \midrule
GCN & 0.877±1.063 & 0.856±0.399 & 101.444±308.123 & 4.823±9.320 & 42.921±0.993 & 6.551±0.075 \\
DI-GCN & 0.451±0.015 & 0.671±0.011 & 1.412±0.012 & 1.188±0.005 & 46.374±0.590 & 6.809±0.043 \\
BI-GCN & 0.270±0.036 & 0.519±0.033 & 21.490±58.748 & 2.861±3.844 & 30.004±2.192 & 5.474±0.201 \\
GIN & 0.337±0.041 & 0.580±0.034 & 1.436±0.053 & 1.198±0.022 & 34.398±11.114 & 5.806±0.852 \\
DI-GIN & 0.356±0.071 & 0.594±0.055 & 1.379±0.015 & 1.174±0.006 & 44.154±12.707 & 6.584±0.941 \\
BI-GIN & 0.293±0.026 & 0.541±0.024 & 1.419±0.046 & 1.191±0.019 & 25.822±1.977 & 5.078±0.190 \\
GAT & 0.330±0.032 & 0.574±0.027 & 567.911±1750.531 & 9.859±22.869 & 30.155±1.879 & 5.489±0.169 \\
DI-GAT & 0.412±0.035 & 0.641±0.027 & 1.406±0.078 & 1.185±0.032 & 41.750±10.476 & 6.421±0.753 \\
BI-GAT & 0.198±0.012 & 0.445±0.014 & 1.348±0.052 & 1.160±0.022 & 41.008±13.522 & 6.336±0.976 \\
GPS-T & 0.508±0.059 & 0.712±0.041 & 1.415±0.061 & 1.189±0.025 & 21.815±1.973 & 4.666±0.206 \\
DI-GPS-T & 0.301±0.012 & 0.549±0.011 & 1.447±0.037 & 1.202±0.015 & 22.161±1.355 & 4.705±0.144 \\
BI-GPS-T & 0.314±0.030 & 0.560±0.027 & 1.315±0.050 & 1.146±0.021 & 26.607±10.277 & 5.087±0.897 \\ \midrule
DGCN & 0.772±0.046 & 0.878±0.025 & 1.364±0.000 & 1.168±0.000 & 69.019±1.345 & 8.307±0.081 \\
DiGCN & 0.345±0.014 & 0.587±0.012 & 1.377±0.021 & 1.173±0.009 & 59.337±0.659 & 7.703±0.042 \\
MagNet & 0.285±0.065 & 0.531±0.059 & 1.608±0.127 & 1.267±0.051 & 36.505±2.749 & 6.038±0.227 \\ \bottomrule
\end{tabular}}
\captionof{table}{OOD performance on the AMP dataset.}
\label{tab:amp_ood}
\end{center}

\begin{center}
\resizebox{0.9\textwidth}{!}{\begin{tabular}{@{}ccccccc@{}}
\toprule
\multirow{2}{*}{Method} & \multicolumn{2}{c}{DSP} & \multicolumn{2}{c}{LUT} & \multicolumn{2}{c}{CP} \\ \cmidrule(l){2-7} 
 & mse & R2 & mse & R2 & mse & R2 \\ \midrule
GCN & 10.817±0.284 & 0.878±0.006 & 0.440±0.084 & 0.879±0.022 & 0.559±0.033 & 0.784±0.014 \\
DI-GCN & 11.101±0.468 & 0.873±0.006 & 0.454±0.128 & 0.863±0.053 & 0.593±0.009 & 0.774±0.006 \\
BI-GCN & 9.996±0.319 & 0.884±0.007 & 0.627±0.158 & 0.857±0.026 & 0.514±0.019 & 0.803±0.013 \\
GINE & 3.720±0.154 & 0.961±0.001 & 0.129±0.042 & 0.966±0.009 & 0.491±0.017 & 0.809±0.008 \\
DI-GINE & 3.194±0.122 & 0.967±0.001 & 0.118±0.022 & 0.968±0.005 & 0.513±0.025 & 0.804±0.015 \\
BI-GINE & 3.244±0.102 & 0.966±0.001 & 0.110±0.029 & 0.971±0.007 & 0.476±0.027 & 0.815±0.011 \\
GAT & 5.440±0.271 & 0.941±0.003 & 0.368±0.054 & 0.900±0.015 & 0.496±0.019 & 0.809±0.007 \\
DI-GAT & 8.927±0.355 & 0.895±0.005 & 0.456±0.135 & 0.856±0.040 & 0.551±0.034 & 0.784±0.015 \\
BI-GAT & 5.418±0.305 & 0.942±0.003 & 0.218±0.054 & 0.938±0.023 & 0.466±0.026 & 0.821±0.013 \\
GPS & 3.343±0.147 & 0.966±0.001 & 0.145±0.026 & 0.962±0.007 & 0.543±0.124 & 0.793±0.024 \\
DI-GPS & 3.210±0.146 & 0.967±0.001 & 0.139±0.032 & 0.964±0.008 & 0.461±0.016 & 0.820±0.014 \\
BI-GPS & 3.209±0.263 & 0.967±0.001 & 0.133±0.027 & 0.968±0.006 & 0.496±0.017 & 0.812±0.016 \\ \midrule
DGCN & 20.220±1.474 & 0.756±0.024 & 0.647±0.228 & 0.797±0.095 & 1.159±0.093 & 0.519±0.028 \\
DiGCN & 10.922±0.353 & 0.880±0.004 & 0.665±0.257 & 0.758±0.098 & 0.613±0.022 & 0.770±0.011 \\
MagNet & 5.048±0.499 & 0.947±0.004 & 0.168±0.049 & 0.955±0.012 & 0.557±0.053 & 0.793±0.024 \\ \bottomrule
\end{tabular}}
\captionof{table}{OOD performance on the HLS dataset.}
\label{tab:hls_ood}
\end{center}

\clearpage

\vspace*{-1.2cm}
\begin{center}
\resizebox{0.9\textwidth}{!}{\begin{tabular}{@{}ccccccccc@{}}
\toprule
Method & blabla & usb\_cdc\_core & wbqspiflash & cic\_decimator & picorv32a & zipdiv & usb & average \\ \midrule
GCN & 9.263±1.994 & 2.888±0.576 & 0.382±0.266 & 0.684±0.151 & 8.796±1.773 & 8.796±1.773 & 0.501±0.076 & 3.278±0.318 \\
DI-GCN & 13.008±5.156 & 2.764±0.347 & 0.504±0.245 & 0.594±0.293 & 10.166±2.557 & 10.166±2.557 & 0.622±0.087 & 4.058±0.892 \\
BI-GCN & 0.731±0.143 & 1.308±0.559 & 0.091±0.044 & 0.069±0.027 & 1.447±0.198 & 1.447±0.198 & 0.093±0.058 & 0.575±0.108 \\
GINE & 0.183±0.023 & 0.073±0.014 & 0.027±0.005 & 0.009±0.003 & 0.549±0.033 & 0.549±0.033 & 0.024±0.005 & 0.125±0.008 \\
DI-GINE & 1.133±0.156 & 0.166±0.041 & 0.147±0.041 & 0.048±0.024 & 1.361±0.398 & 1.361±0.398 & 0.123±0.051 & 0.439±0.048 \\
BI-GINE & 0.142±0.014 & 0.111±0.083 & 0.023±0.006 & 0.010±0.005 & 0.574±0.041 & 0.574±0.041 & 0.032±0.013 & 0.132±0.012 \\
GAT & 17.032±16.149 & 3.128±0.521 & 0.477±0.603 & 0.393±0.219 & 6.910±3.546 & 6.910±3.546 & 0.448±0.296 & 4.204±3.054 \\
\bottomrule
\end{tabular}}
\vspace{-0.2cm}
\captionof{table}{ID performance with `mse' metric on the Time dataset to predict hold slack.}
\label{tab:time_hold_id_mse}
\end{center}
\vspace{-0.6cm}
\begin{center}
\resizebox{0.9\textwidth}{!}{\begin{tabular}{@{}ccccccccc@{}}
\toprule
Method & blabla & usb\_cdc\_core & wbqspiflash & cic\_decimator & picorv32a & zipdiv & usb & average \\ \midrule
GCN & 0.910±0.019 & -0.312±0.262 & 0.826±0.120 & 0.358±0.142 & 0.460±0.108 & 0.905±0.029 & 0.539±0.070 & 0.526±0.045 \\
DI-GCN & 0.873±0.050 & -0.255±0.157 & 0.771±0.110 & 0.442±0.275 & 0.376±0.156 & 0.835±0.070 & 0.428±0.079 & 0.496±0.056 \\
BI-GCN & 0.992±0.001 & 0.405±0.254 & 0.958±0.020 & 0.935±0.025 & 0.911±0.012 & 0.937±0.011 & 0.914±0.053 & 0.865±0.047 \\
GINE & 0.998±0.000 & 0.966±0.006 & 0.987±0.002 & 0.991±0.003 & 0.966±0.002 & 0.997±0.001 & 0.977±0.004 & 0.983±0.001 \\
DI-GINE & 0.988±0.001 & 0.924±0.018 & 0.933±0.018 & 0.954±0.022 & 0.916±0.024 & 0.978±0.002 & 0.886±0.047 & 0.940±0.007 \\
BI-GINE & 0.998±0.000 & 0.949±0.037 & 0.989±0.002 & 0.990±0.005 & 0.964±0.002 & 0.993±0.005 & 0.970±0.012 & 0.979±0.006 \\
GAT & 0.834±0.156 & -0.421±0.236 & 0.784±0.272 & 0.630±0.205 & 0.575±0.217 & 0.772±0.230 & 0.588±0.271 & 0.537±0.168 \\
\bottomrule
\end{tabular}}
\vspace{-0.2cm}
\captionof{table}{ID performance with `R2' metric on the Time dataset to predict hold slack.}
\label{tab:time_id_hold_r2}
\end{center}
\vspace{-0.6cm}
\begin{center}
\resizebox{0.9\textwidth}{!}{\begin{tabular}{@{}ccccccccc@{}}
\toprule
Method & blabla & usb\_cdc\_core & wbqspiflash & cic\_decimator & picorv32a & zipdiv & usb & average \\ \midrule
GCN & 127.555±9.301 & 17.551±1.646 & 34.125±1.436 & 1.399±0.128 & 126.288±10.077 & 28.960±1.970 & 17.849±1.509 & 50.532±1.053 \\
DI-GCN & 99.699±13.849 & 18.302±5.510 & 30.164±2.874 & 2.984±4.527 & 135.526±16.281 & 26.859±5.355 & 15.223±4.432 & 46.965±2.204 \\
BI-GCN & 104.062±11.187 & 12.867±1.746 & 27.941±1.197 & 4.693±0.572 & 102.680±9.184 & 20.688±1.020 & 6.033±2.329 & 39.852±2.405 \\
GINE & 24.911±9.848 & 2.819±0.801 & 8.012±3.029 & 1.233±0.360 & 31.441±2.463 & 3.379±1.141 & 2.211±0.517 & 10.572±2.027 \\
DI-GINE & 25.642±6.938 & 2.207±0.586 & 7.105±2.835 & 1.823±0.632 & 21.690±1.080 & 4.088±1.363 & 1.785±0.315 & 9.191±1.564 \\
BI-GINE & 21.108±6.326 & 1.690±0.510 & 4.484±1.648 & 1.079±0.778 & 23.207±5.134 & 3.160±2.036 & 2.695±2.925 & 8.203±1.720 \\
GAT & 105.449±39.675 & 13.554±3.935 & 32.138±2.692 & 3.416±1.475 & 116.730±16.637 & 25.274±2.653 & 7.188±2.344 & 43.393±3.497 \\
\bottomrule
\end{tabular}}
\vspace{-0.2cm}
\captionof{table}{ID performance with `R2' metric on the Time dataset to predict setup slack.}
\label{tab:time_setup_id_mse}
\end{center}
\vspace{-0.6cm}
\begin{center}
\resizebox{0.9\textwidth}{!}{\begin{tabular}{@{}ccccccccc@{}}
\toprule
Method & blabla & usb\_cdc\_core & wbqspiflash & cic\_decimator & picorv32a & zipdiv & usb & average \\ \midrule
GCN & -1.030±0.148 & -4.397±0.506 & -0.219±0.051 & 0.180±0.075 & -2.489±0.278 & -0.971±0.134 & -4.437±0.459 & -1.909±0.150 \\
DI-GCN & -0.586±0.220 & -4.628±1.694 & -0.077±0.102 & -0.747±2.650 & -2.744±0.449 & -0.828±0.364 & -3.637±1.350 & -1.892±0.254 \\
BI-GCN & -0.656±0.178 & -2.956±0.537 & 0.001±0.042 & -1.747±0.335 & -1.837±0.253 & -0.408±0.069 & -0.837±0.709 & -1.206±0.144 \\
GINE & 0.603±0.156 & 0.132±0.246 & 0.713±0.108 & 0.278±0.211 & 0.131±0.068 & 0.769±0.077 & 0.326±0.157 & 0.422±0.078 \\
DI-GINE & 0.591±0.110 & 0.321±0.180 & 0.746±0.101 & -0.067±0.370 & 0.400±0.029 & 0.721±0.092 & 0.456±0.096 & 0.452±0.055 \\
BI-GINE & 0.664±0.100 & 0.480±0.157 & 0.839±0.058 & 0.367±0.455 & 0.358±0.141 & 0.784±0.138 & 0.178±0.891 & 0.524±0.183 \\
GAT & -0.678±0.631 & -3.168±1.210 & -0.148±0.096 & -1.000±0.863 & -2.225±0.459 & -0.720±0.180 & -1.189±0.714 & -1.304±0.175 \\
\bottomrule
\end{tabular}}
\vspace{-0.2cm}
\captionof{table}{ID performance with `R2' metric on the Time dataset to predict setup slack.}
\label{tab:time_setup_id_r2}
\end{center}
\vspace{-0.6cm}
\begin{center}
\resizebox{0.9\textwidth}{!}{
\begin{tabular}{@{}ccccccccc@{}}
\toprule
 & \multicolumn{4}{c}{hold} & \multicolumn{4}{c}{setup} \\ \midrule
\multirow{2}{*}{Method} & \multicolumn{2}{c}{xtea} & \multicolumn{2}{c}{synth\_ram} & \multicolumn{2}{c}{xtea} & \multicolumn{2}{c}{synth\_ram} \\
 & mse & r2 & mse & r2 & mse & r2 & mse & r2 \\ \midrule
GCN & 6.745±0.853 & -0.062±0.134 & 7.511±1.387 & -1.030±0.375 & 117.599±4.681 & -2.963±0.157 & 509.769±16.052 & -984.202±31.023 \\
DI-GCN & 7.010±0.592 & -0.104±0.093 & 9.764±2.355 & -1.640±0.636 & 116.332±16.048 & -2.920±0.540 & 418.819±24.262 & -808.427±46.890 \\
BI-GCN & 5.675±0.784 & 0.105±0.123 & 2.446±0.255 & 0.338±0.069 & 100.915±4.158 & -2.401±0.140 & 371.644±10.433 & -717.255±20.164 \\
GINE & 4.446±1.623 & 0.299±0.255 & 2.482±1.650 & 0.328±0.446 & 62.927±28.904 & -1.120±0.974 & 1341.500±430.261 & -2591.639±831.541 \\
DI-GINE & 3.075±0.887 & 0.515±0.139 & 9.545±4.222 & -1.581±1.141 & 61.694±6.981 & -1.079±0.235 & 612.807±101.175 & -1183.337±195.535 \\
\bottomrule
\end{tabular}
}
\vspace{-0.2cm}
\captionof{table}{OOD performance on the TIME dataset.}
\label{tab:time_ood}
\end{center}
\vspace{-0.8cm}

\clearpage
\vspace*{-2.2cm}
\begin{center}
\resizebox{0.8\textwidth}{!}{\begin{tabular}{@{}ccccccc@{}}
\toprule
Method & densenets & mnasnets & mobilenetv2s & mobilenetv3s & nasbench201s & average \\ \midrule
GCN & 43.042±1.663 & 7.764±0.775 & 30.222±0.817 & 5.981±1.454 & 2.134±0.772 & 17.828±0.802 \\
DI-GCN & 41.241±0.336 & 6.704±0.183 & 29.457±0.389 & 5.275±0.262 & 1.129±0.426 & 16.761±0.151 \\
BI-GCN & 41.271±1.729 & 8.054±0.399 & 29.473±0.399 & 5.365±0.159 & 2.114±0.731 & 17.256±0.238 \\
GIN & 7.694±0.442 & 1.833±0.187 & 3.535±0.343 & 1.491±0.343 & 0.617±0.216 & 3.034±0.172 \\
DI-GIN & 9.894±0.450 & 1.943±0.116 & 3.395±0.293 & 1.157±0.118 & 0.748±0.118 & 3.427±0.129 \\
BI-GIN & 7.615±0.247 & 1.901±0.151 & 3.339±0.172 & 1.177±0.160 & 0.493±0.069 & 2.905±0.098 \\
GAT & 14.272±4.836 & 4.160±0.460 & 8.733±0.936 & 3.227±0.456 & 0.577±0.071 & 6.194±1.242 \\
DI-GAT & 21.741±2.897 & 3.850±0.418 & 11.801±1.780 & 3.640±0.582 & 1.820±0.159 & 8.571±0.900 \\
BI-GAT & 9.322±0.535 & 2.371±0.469 & 4.957±0.497 & 1.723±0.481 & 0.498±0.086 & 3.774±0.262 \\
GPS-P & 7.080±0.302 & 2.125±0.294 & 3.951±0.289 & 1.692±0.395 & 0.723±0.197 & 3.114±0.132 \\
DI-GPS-P & 6.821±0.223 & 2.016±0.226 & 3.265±0.494 & 1.337±0.618 & 0.620±0.372 & 2.812±0.257 \\
BI-GPS-P & 6.863±0.265 & 1.971±0.222 & 3.794±0.311 & 1.399±0.361 & 0.618±0.260 & 2.929±0.149 \\ \midrule
DGCN & 41.386±0.878 & 6.651±0.253 & 30.312±0.307 & 4.591±0.310 & 2.101±0.211 & 17.008±0.152 \\
DiGCN & 42.506±0.745 & 7.203±0.137 & 30.941±0.426 & 5.263±0.468 & 1.879±0.558 & 17.558±0.169 \\
MagNet & 7.368±0.189 & 1.910±0.152 & 3.015±0.280 & 1.290±0.315 & 0.514±0.166 & 2.819±0.177 \\ \midrule
nn-meter & 7.1 & 3.19 & 3.25 & 2.03 & 0.44 & 3.20 \\ \bottomrule
\end{tabular}}
\vspace{-0.2cm}
\captionof{table}{ID performance on the CG dataset on device `Cortex A76 CPU' with `rmse' metric.}
\label{tab:cpu_id_rmse}
\end{center}
\vspace{-0.5cm}
\begin{center}
\resizebox{0.8\textwidth}{!}{\begin{tabular}{@{}ccccccc@{}}
\toprule
Method & densenets & mnasnets & mobilenetv3s & mobilenetv4s & nasbench202s & average \\ \midrule
GCN & 0.217±0.034 & 0.267±0.033 & 0.118±0.026 & 0.239±0.090 & 0.290±0.183 & 0.226±0.037 \\
DI-GCN & 0.222±0.010 & 0.289±0.022 & 0.101±0.014 & 0.264±0.023 & 0.530±0.184 & 0.281±0.037 \\
BI-GCN & 0.245±0.038 & 0.269±0.012 & 0.111±0.022 & 0.287±0.010 & 0.272±0.152 & 0.236±0.031 \\
GIN & 0.876±0.028 & 0.863±0.054 & 0.680±0.050 & 0.723±0.078 & 0.716±0.190 & 0.771±0.052 \\
DI-GIN & 0.764±0.025 & 0.835±0.040 & 0.693±0.050 & 0.805±0.060 & 0.626±0.114 & 0.744±0.021 \\
BI-GIN & 0.892±0.020 & 0.842±0.047 & 0.681±0.033 & 0.861±0.053 & 0.858±0.059 & 0.826±0.024 \\
GAT & 0.662±0.107 & 0.456±0.067 & 0.359±0.032 & 0.476±0.060 & 0.786±0.056 & 0.547±0.050 \\
DI-GAT & 0.466±0.053 & 0.540±0.062 & 0.273±0.046 & 0.375±0.046 & 0.343±0.077 & 0.399±0.029 \\
BI-GAT & 0.808±0.034 & 0.746±0.077 & 0.511±0.065 & 0.689±0.116 & 0.844±0.091 & 0.719±0.051 \\
GPS-P & 0.904±0.019 & 0.809±0.051 & 0.650±0.041 & 0.666±0.126 & 0.697±0.173 & 0.745±0.057 \\
DI-GPS-P & 0.933±0.018 & 0.831±0.071 & 0.735±0.038 & 0.796±0.234 & 0.750±0.271 & 0.808±0.104 \\
BI-GPS-P & 0.922±0.020 & 0.836±0.048 & 0.666±0.051 & 0.738±0.141 & 0.750±0.215 & 0.782±0.072 \\ \midrule
DGCN & 0.188±0.017 & 0.320±0.019 & 0.105±0.022 & 0.294±0.042 & 0.269±0.042 & 0.235±0.009 \\
DiGCN & 0.199±0.025 & 0.289±0.035 & 0.085±0.014 & 0.288±0.042 & 0.309±0.111 & 0.233±0.017 \\
MagNet & 0.896±0.012 & 0.835±0.060 & 0.730±0.048 & 0.830±0.113 & 0.851±0.108 & 0.828±0.062 \\ \midrule
nn-meter & 0.931 & 0.824 & 0.676 & 0.738 & 0.824 & 0.798 \\ \bottomrule
\end{tabular}}
\vspace{-0.2cm}
\captionof{table}{ID performance on the CG dataset on device `Cortex A76 CPU' with `acc5' metric.}
\label{tab:cg_cpu_id_acc5}
\end{center}
\vspace{-0.5cm}
\begin{center}
\resizebox{0.8\textwidth}{!}{\begin{tabular}{@{}ccccccc@{}}
\toprule
Method & densenets & mnasnets & mobilenetv4s & mobilenetv5s & nasbench203s & average \\ \midrule
GCN & 0.434±0.041 & 0.511±0.042 & 0.222±0.031 & 0.453±0.122 & 0.515±0.233 & 0.426±0.053 \\
DI-GCN & 0.446±0.023 & 0.571±0.023 & 0.193±0.021 & 0.476±0.032 & 0.818±0.158 & 0.501±0.031 \\
BI-GCN & 0.481±0.025 & 0.483±0.026 & 0.243±0.018 & 0.536±0.029 & 0.517±0.199 & 0.451±0.037 \\
GIN & 0.998±0.004 & 0.997±0.004 & 0.936±0.024 & 0.941±0.044 & 0.972±0.035 & 0.968±0.016 \\
DI-GIN & 0.984±0.013 & 0.999±0.003 & 0.965±0.019 & 0.982±0.018 & 0.956±0.025 & 0.977±0.008 \\
BI-GIN & 1.000±0.000 & 0.996±0.005 & 0.952±0.018 & 0.991±0.017 & 0.994±0.012 & 0.986±0.007 \\
GAT & 0.924±0.091 & 0.808±0.039 & 0.645±0.063 & 0.730±0.060 & 0.973±0.017 & 0.816±0.036 \\
DI-GAT & 0.757±0.055 & 0.828±0.054 & 0.532±0.050 & 0.650±0.075 & 0.625±0.071 & 0.678±0.036 \\
BI-GAT & 0.993±0.009 & 0.977±0.034 & 0.821±0.057 & 0.937±0.081 & 0.994±0.007 & 0.944±0.026 \\
GPS-P & 1.000±0.000 & 0.996±0.006 & 0.918±0.033 & 0.901±0.080 & 0.928±0.085 & 0.949±0.034 \\
DI-GPS-P & 0.999±0.003 & 0.997±0.004 & 0.974±0.018 & 0.949±0.117 & 0.940±0.154 & 0.971±0.055 \\
BI-GPS-P & 0.999±0.003 & 0.998±0.004 & 0.938±0.021 & 0.922±0.085 & 0.953±0.078 & 0.962±0.030 \\ \cmidrule(r){1-4} \cmidrule(l){6-7} 
DGCN & 0.409±0.015 & 0.568±0.021 & 0.204±0.020 & 0.538±0.027 & 0.504±0.076 & 0.444±0.010 \\
DiGCN & 0.390±0.017 & 0.543±0.017 & 0.186±0.022 & 0.522±0.044 & 0.582±0.156 & 0.444±0.029 \\
MagNet & 1.000±0.000 & 0.996±0.005 & 0.973±0.028 & 0.983±0.031 & 0.988±0.019 & 0.988±0.015 \\ \midrule
nn-meter & 0.999 & 0.992 & 0.977 & 0.990 & 0.999 & 0.991 \\ \bottomrule
\end{tabular}}
\vspace{-0.2cm}
\captionof{table}{ID performance on the CG dataset on device `Cortex A76 CPU' with `acc10' metric.}
\label{tab:cg_cpu_id_acc10}
\end{center}
\vspace{-0.5cm}
\begin{center}
\resizebox{0.8\textwidth}{!}{\begin{tabular}{@{}ccccccc@{}}
\toprule
\multirow{2}{*}{Method} & \multicolumn{3}{c}{proxylessnass} & \multicolumn{3}{c}{resnets} \\ \cmidrule(l){2-7} 
 & rmse & acc5 & acc10 & rmse & acc5 & acc10 \\ \midrule
GCN & 38.016±9.318 & 0.130±0.018 & 0.237±0.040 & 529.239±19.212 & 0.012±0.007 & 0.017±0.009 \\
DI-GCN & 33.581±1.016 & 0.143±0.019 & 0.264±0.016 & 517.553±4.419 & 0.010±0.002 & 0.020±0.003 \\
BI-GCN & 38.612±2.911 & 0.112±0.010 & 0.224±0.007 & 524.188±11.204 & 0.008±0.009 & 0.016±0.011 \\
GIN & 18.768±4.710 & 0.156±0.069 & 0.333±0.123 & 326.443±37.550 & 0.059±0.024 & 0.112±0.038 \\
DI-GIN & 10.742±0.876 & 0.306±0.021 & 0.566±0.051 & 455.235±16.044 & 0.032±0.010 & 0.073±0.012 \\
BI-GIN & 11.097±2.002 & 0.329±0.061 & 0.589±0.096 & 362.930±29.676 & 0.046±0.015 & 0.087±0.013 \\
GAT & 20.289±4.934 & 0.208±0.059 & 0.396±0.090 & 467.151±29.254 & 0.012±0.017 & 0.025±0.026 \\
DI-GAT & 19.306±3.641 & 0.192±0.043 & 0.386±0.063 & 386.181±44.349 & 0.033±0.023 & 0.069±0.030 \\
BI-GAT & 14.833±4.066 & 0.237±0.070 & 0.471±0.132 & 472.310±24.953 & 0.019±0.011 & 0.036±0.014 \\
GPS-P & 11.952±2.043 & 0.275±0.078 & 0.527±0.104 & 473.207±15.942 & 0.013±0.007 & 0.023±0.011 \\
DI-GPS-P & 10.122±0.911 & 0.293±0.038 & 0.588±0.075 & 490.252±11.336 & 0.005±0.005 & 0.012±0.009 \\
BI-GPS-P & 12.188±1.565 & 0.249±0.054 & 0.506±0.074 & 475.745±14.259 & 0.005±0.009 & 0.015±0.012 \\
DGCN & 28.038±1.707 & 0.123±0.016 & 0.274±0.015 & 535.961±5.274 & 0.003±0.005 & 0.007±0.007 \\
DiGCN & 26.308±1.739 & 0.138±0.017 & 0.287±0.022 & 542.416±5.718 & 0.001±0.002 & 0.005±0.004 \\
MagNet & 9.282±1.321 & 0.433±0.069 & 0.725±0.055 & 483.296±9.749 & 0.015±0.010 & 0.024±0.017 \\ \midrule
nn-meter & 3.18 & 0.846 & 1.00 & 7.19 & 0.845 & 0.999 \\ \bottomrule
\end{tabular}}
\vspace{-0.2cm}
\captionof{table}{OOD performance on the CG dataset on device `Cortex A76 CPU'.}
\label{tab:cg_cpu_ood}
\end{center}

\clearpage
\vspace{-1.2cm}
\begin{center}
\resizebox{0.8\textwidth}{!}{\begin{tabular}{cllllll}
\hline
Method & \multicolumn{1}{c}{densenets} & \multicolumn{1}{c}{mnasnets} & \multicolumn{1}{c}{mobilenetv3s} & \multicolumn{1}{c}{mobilenetv4s} & \multicolumn{1}{c}{nasbench202s} & \multicolumn{1}{c}{average} \\ \hline
GCN & 6.330±0.099 & 1.192±0.035 & 5.075±0.054 & 0.811±0.034 & 0.110±0.029 & 2.704±0.016 \\
DI-GCN & 6.226±0.061 & 1.165±0.029 & 5.088±0.035 & 0.795±0.032 & 0.128±0.036 & 2.681±0.013 \\
BI-GCN & 6.233±0.140 & 1.173±0.094 & 4.091±0.158 & 0.867±0.098 & 0.411±0.118 & 2.555±0.052 \\
GIN & 0.514±0.056 & 0.133±0.021 & 0.322±0.064 & 0.110±0.017 & 0.079±0.020 & 0.231±0.025 \\
DI-GIN & 0.909±0.064 & 0.176±0.017 & 0.385±0.036 & 0.156±0.029 & 0.250±0.022 & 0.375±0.019 \\
BI-GIN & 0.554±0.038 & 0.158±0.016 & 0.357±0.037 & 0.142±0.021 & 0.116±0.033 & 0.265±0.013 \\
GAT & 1.742±0.302 & 0.446±0.079 & 1.329±0.245 & 0.349±0.081 & 0.132±0.050 & 0.800±0.102 \\
DI-GAT & 1.904±0.173 & 0.531±0.087 & 1.810±0.414 & 0.481±0.068 & 0.497±0.022 & 1.044±0.133 \\
BI-GAT & 0.854±0.057 & 0.192±0.048 & 0.510±0.094 & 0.169±0.035 & 0.080±0.020 & 0.361±0.039 \\
GPS-P & 0.313±0.021 & 0.131±0.015 & 0.286±0.035 & 0.104±0.013 & 0.069±0.025 & 0.181±0.009 \\
DI-GPS-P & 0.320±0.020 & 0.149±0.019 & 0.321±0.081 & 0.147±0.029 & 0.186±0.134 & 0.225±0.032 \\
BI-GPS-P & 0.486±0.062 & 0.197±0.039 & 0.508±0.117 & 0.174±0.047 & 0.124±0.072 & 0.298±0.054 \\ \hline
DGCN & 6.695±0.428 & 1.267±0.101 & 5.396±0.050 & 0.817±0.200 & 0.614±0.079 & 2.958±0.074 \\
DiiGCN & 6.298±0.165 & 1.344±0.106 & 5.508±0.185 & 0.919±0.107 & 0.549±0.174 & 2.924±0.083 \\
MagNet & 0.478±0.032 & 0.120±0.015 & 0.230±0.017 & 0.106±0.008 & 0.060±0.008 & 0.199±0.008 \\ \hline
\end{tabular}}
\vspace{-0.2cm}
\captionof{table}{ID performance on the CG dataset on device `Adreno 630 GPU' with `rmse' metric.}
\label{tab:cg_gpu_id_rmse}
\end{center}
\vspace{-0.5cm}
\begin{center}
\resizebox{0.8\textwidth}{!}{\begin{tabular}{cllllll}
\hline
Method & \multicolumn{1}{c}{densenets} & \multicolumn{1}{c}{mnasnets} & \multicolumn{1}{c}{mobilenetv3s} & \multicolumn{1}{c}{mobilenetv4s} & \multicolumn{1}{c}{nasbench202s} & \multicolumn{1}{c}{average} \\ \hline
GCN & 0.325±0.032 & 0.383±0.019 & 0.119±0.015 & 0.410±0.035 & 0.943±0.052 & 0.435±0.013 \\
DI-GCN & 0.351±0.014 & 0.404±0.023 & 0.127±0.031 & 0.442±0.047 & 0.910±0.076 & 0.447±0.020 \\
BI-GCN & 0.354±0.030 & 0.410±0.040 & 0.164±0.031 & 0.424±0.038 & 0.485±0.100 & 0.367±0.028 \\
GIN & 1.000±0.000 & 1.000±0.000 & 0.956±0.032 & 1.000±0.000 & 0.994±0.009 & 0.990±0.007 \\
DI-GIN & 0.986±0.014 & 1.000±0.000 & 0.916±0.023 & 0.997±0.006 & 0.703±0.062 & 0.920±0.009 \\
BI-GIN & 1.000±0.000 & 1.000±0.000 & 0.917±0.016 & 0.991±0.012 & 0.945±0.057 & 0.970±0.012 \\
GAT & 0.872±0.054 & 0.864±0.068 & 0.515±0.070 & 0.862±0.065 & 0.907±0.131 & 0.804±0.050 \\
DI-GAT & 0.839±0.049 & 0.790±0.073 & 0.344±0.085 & 0.768±0.047 & 0.466±0.052 & 0.641±0.046 \\
BI-GAT & 0.989±0.005 & 0.992±0.013 & 0.829±0.049 & 0.993±0.012 & 0.994±0.009 & 0.959±0.011 \\
GPS-P & 1.000±0.000 & 1.000±0.000 & 0.966±0.021 & 1.000±0.000 & 0.994±0.015 & 0.992±0.005 \\
DI-GPS-P & 1.000±0.000 & 1.000±0.000 & 0.927±0.064 & 0.976±0.049 & 0.749±0.351 & 0.930±0.068 \\
BI-GPS-P & 1.000±0.000 & 0.995±0.005 & 0.861±0.043 & 0.972±0.043 & 0.892±0.156 & 0.944±0.045 \\ \hline
DGCN & 0.290±0.019 & 0.403±0.037 & 0.114±0.020 & 0.471±0.113 & 0.302±0.074 & 0.316±0.038 \\
DiiGCN & 0.302±0.028 & 0.432±0.042 & 0.118±0.022 & 0.441±0.054 & 0.363±0.146 & 0.331±0.031 \\
MagNet & 1.000±0.000 & 1.000±0.000 & 0.978±0.011 & 1.000±0.000 & 0.999±0.003 & 0.995±0.002 \\ \hline
\end{tabular}}
\vspace{-0.2cm}
\captionof{table}{ID performance on the CG dataset on device `Adreno 630 GPU' with `acc5' metric.}
\label{tab:cg_gpu630_id_acc5}
\end{center}
\vspace{-0.5cm}
\begin{center}
\resizebox{0.8\textwidth}{!}{\begin{tabular}{cllllll}
\hline
Method & \multicolumn{1}{c}{densenets} & \multicolumn{1}{c}{mnasnets} & \multicolumn{1}{c}{mobilenetv3s} & \multicolumn{1}{c}{mobilenetv4s} & \multicolumn{1}{c}{nasbench202s} & \multicolumn{1}{c}{average} \\ \hline
GCN & 0.608±0.017 & 0.716±0.022 & 0.227±0.018 & 0.780±0.048 & 0.998±0.004 & 0.665±0.009 \\
DI-GCN & 0.614±0.014 & 0.741±0.021 & 0.225±0.015 & 0.792±0.048 & 0.992±0.013 & 0.672±0.013 \\
BI-GCN & 0.629±0.028 & 0.772±0.048 & 0.312±0.029 & 0.747±0.086 & 0.789±0.088 & 0.650±0.037 \\
GIN & 1.000±0.000 & 1.000±0.000 & 0.994±0.007 & 1.000±0.000 & 1.000±0.000 & 0.999±0.001 \\
DI-GIN & 1.000±0.000 & 1.000±0.000 & 0.994±0.005 & 1.000±0.000 & 0.904±0.010 & 0.979±0.002 \\
BI-GIN & 1.000±0.000 & 1.000±0.000 & 0.993±0.006 & 1.000±0.000 & 0.980±0.024 & 0.994±0.004 \\
GAT & 0.989±0.017 & 0.992±0.013 & 0.802±0.079 & 0.989±0.018 & 0.993±0.018 & 0.953±0.024 \\
DI-GAT & 0.991±0.005 & 0.980±0.020 & 0.630±0.106 & 0.962±0.014 & 0.715±0.033 & 0.855±0.029 \\
BI-GAT & 1.000±0.000 & 1.000±0.000 & 0.985±0.012 & 1.000±0.000 & 1.000±0.000 & 0.997±0.002 \\
GPS-P & 1.000±0.000 & 1.000±0.000 & 0.998±0.006 & 1.000±0.000 & 1.000±0.000 & 0.999±0.001 \\
DI-GPS-P & 1.000±0.000 & 1.000±0.000 & 0.988±0.021 & 1.000±0.000 & 0.947±0.157 & 0.986±0.030 \\
BI-GPS-P & 1.000±0.000 & 1.000±0.000 & 0.990±0.010 & 0.998±0.003 & 0.961±0.059 & 0.990±0.012 \\ \hline
DGCN & 0.564±0.022 & 0.708±0.031 & 0.200±0.016 & 0.781±0.103 & 0.611±0.099 & 0.573±0.041 \\
DiiGCN & 0.597±0.017 & 0.721±0.031 & 0.220±0.033 & 0.724±0.088 & 0.613±0.197 & 0.575±0.054 \\
MagNet & 1.000±0.000 & 1.000±0.000 & 1.000±0.000 & 1.000±0.000 & 1.000±0.000 & 1.000±0.000 \\ \hline
\end{tabular}}
\vspace{-0.2cm}
\captionof{table}{ID performance on the CG dataset on device `Adreno 630 GPU' with `acc10' metric.}
\label{tab:cg_gpu630_id_acc10}
\end{center}
\vspace{-0.5cm}
\begin{center}
\resizebox{0.8\textwidth}{!}{\begin{tabular}{ccccccc}
\hline
\multirow{2}{*}{Method} & \multicolumn{3}{c}{proxylessnass} & \multicolumn{3}{c}{resnets} \\ \cline{2-7} 
 & rmse & acc5 & acc10 & rmse & acc5 & acc10 \\ \hline
GCN & 6.517±0.386 & 0.150±0.019 & 0.275±0.017 & 111.019±1.367 & 0.006±0.005 & 0.015±0.007 \\
DI-GCN & 6.135±0.149 & 0.154±0.021 & 0.305±0.014 & 111.631±1.095 & 0.007±0.005 & 0.012±0.006 \\
BI-GCN & 5.314±0.297 & 0.150±0.013 & 0.294±0.023 & 98.142±2.642 & 0.002±0.004 & 0.006±0.003 \\
GIN & 3.252±0.272 & 0.279±0.046 & 0.512±0.024 & 90.353±5.675 & 0.030±0.012 & 0.066±0.017 \\
DI-GIN & 3.121±0.549 & 0.271±0.076 & 0.514±0.079 & 109.263±2.076 & 0.028±0.013 & 0.057±0.011 \\
BI-GIN & 3.307±0.184 & 0.306±0.044 & 0.538±0.020 & 99.362±2.257 & 0.030±0.017 & 0.053±0.020 \\
GAT & 3.082±0.419 & 0.252±0.065 & 0.479±0.084 & 95.114±5.394 & 0.011±0.014 & 0.025±0.030 \\
DI-GAT & 3.982±0.341 & 0.172±0.019 & 0.355±0.027 & 92.212±9.020 & 0.035±0.019 & 0.072±0.029 \\
BI-GAT & 3.461±0.376 & 0.265±0.038 & 0.476±0.059 & 109.787±1.405 & 0.003±0.005 & 0.005±0.009 \\
GPS-P & 2.554±0.146 & 0.304±0.068 & 0.639±0.022 & 104.410±1.863 & 0.010±0.005 & 0.019±0.006 \\
DI-GPS-P & 2.515±0.148 & 0.270±0.070 & 0.614±0.061 & 105.275±2.238 & 0.006±0.010 & 0.009±0.013 \\
BI-GPS-P & 2.793±0.261 & 0.281±0.064 & 0.542±0.091 & 104.383±3.367 & 0.009±0.007 & 0.018±0.011 \\
DGCN & 5.794±0.249 & 0.160±0.012 & 0.293±0.026 & 116.229±1.179 & 0.006±0.006 & 0.017±0.010 \\
DiGCN & 6.042±0.906 & 0.140±0.023 & 0.271±0.043 & 117.250±1.149 & 0.006±0.005 & 0.013±0.008 \\
MagNet & 3.120±0.409 & 0.434±0.077 & 0.599±0.070 & 109.707±2.288 & 0.001±0.002 & 0.003±0.003 \\ \hline
\end{tabular}}
\vspace{-0.2cm}
\captionof{table}{OOD performance on the CG dataset on device `Adreno 630 GPU'.}
\label{tab:cg_gpu630_ood}
\end{center}

\clearpage

\vspace{-1.2cm}
\begin{center}
\resizebox{0.8\textwidth}{!}{\begin{tabular}{@{}ccccccc@{}}
\toprule
Method & densenets & mnasnets & mobilenetv3s & mobilenetv4s & nasbench202s & average \\ \midrule
GCN & 4.754±0.087 & 1.141±0.032 & 4.717±0.039 & 0.820±0.042 & 0.184±0.075 & 2.323±0.026 \\
DI-GCN & 4.899±0.188 & 1.159±0.064 & 4.869±0.179 & 0.834±0.070 & 0.162±0.102 & 2.385±0.084 \\
BI-GCN & 4.712±0.095 & 1.150±0.061 & 3.834±0.104 & 0.887±0.142 & 0.364±0.116 & 2.190±0.042 \\
GIN & 0.847±0.089 & 0.249±0.176 & 0.518±0.161 & 0.216±0.076 & 0.539±0.511 & 0.474±0.103 \\
DI-GIN & 0.639±0.034 & 0.119±0.013 & 0.303±0.019 & 0.102±0.023 & 0.050±0.010 & 0.243±0.011 \\
BI-GIN & 0.310±0.022 & 0.100±0.011 & 0.231±0.015 & 0.099±0.028 & 0.039±0.013 & 0.156±0.011 \\
GAT & 1.630±0.299 & 0.441±0.088 & 1.585±0.311 & 0.388±0.052 & 0.131±0.081 & 0.835±0.127 \\
DI-GAT & 1.763±0.174 & 0.576±0.097 & 1.589±0.162 & 0.513±0.087 & 0.429±0.049 & 0.974±0.077 \\
BI-GAT & 0.830±0.082 & 0.269±0.026 & 0.834±0.114 & 0.264±0.030 & 0.062±0.020 & 0.452±0.032 \\
GPS-P & 0.303±0.035 & 0.132±0.020 & 0.335±0.043 & 0.112±0.026 & 0.106±0.062 & 0.197±0.015 \\
DI-GPS-P & 0.316±0.030 & 0.145±0.020 & 0.332±0.042 & 0.110±0.009 & 0.060±0.032 & 0.193±0.018 \\
BI-GPS-P & 0.296±0.012 & 0.118±0.016 & 0.303±0.049 & 0.083±0.011 & 0.108±0.071 & 0.182±0.026 \\ \midrule
DGCN & 4.903±0.105 & 1.162±0.047 & 5.048±0.030 & 0.756±0.060 & 0.409±0.045 & 2.456±0.032 \\
DiiGCN & 4.807±0.174 & 1.274±0.118 & 5.139±0.174 & 0.834±0.091 & 0.470±0.075 & 2.505±0.094 \\
MagNet & 0.583±0.051 & 0.155±0.019 & 0.337±0.042 & 0.150±0.029 & 0.074±0.031 & 0.260±0.019 \\ \bottomrule
\end{tabular}}
\vspace{-0.2cm}
\captionof{table}{ID performance on the CG dataset on device `Adreno 640 GPU' with `rmse' metric.}
\label{tab:cg_gpu640_id_rmse}
\end{center}
\vspace{-0.5cm}
\begin{center}
\resizebox{0.8\textwidth}{!}{\begin{tabular}{@{}ccccccc@{}}
\toprule
Method & densenets & mnasnets & mobilenetv3s & mobilenetv4s & nasbench202s & average \\ \midrule
GCN & 0.367±0.014 & 0.381±0.024 & 0.133±0.020 & 0.422±0.034 & 0.678±0.320 & 0.396±0.066 \\
DI-GCN & 0.357±0.016 & 0.416±0.035 & 0.139±0.029 & 0.450±0.047 & 0.782±0.272 & 0.429±0.059 \\
BI-GCN & 0.412±0.040 & 0.447±0.038 & 0.178±0.024 & 0.422±0.084 & 0.383±0.180 & 0.368±0.050 \\
GIN & 0.986±0.009 & 0.951±0.124 & 0.806±0.103 & 0.943±0.081 & 0.515±0.409 & 0.840±0.095 \\
DI-GIN & 0.999±0.003 & 1.000±0.000 & 0.962±0.023 & 1.000±0.000 & 0.997±0.006 & 0.991±0.004 \\
BI-GIN & 1.000±0.000 & 1.000±0.000 & 0.983±0.006 & 0.999±0.003 & 1.000±0.000 & 0.996±0.001 \\
GAT & 0.846±0.057 & 0.861±0.075 & 0.427±0.064 & 0.821±0.069 & 0.830±0.287 & 0.757±0.078 \\
DI-GAT & 0.836±0.042 & 0.744±0.080 & 0.373±0.053 & 0.674±0.101 & 0.424±0.058 & 0.610±0.047 \\
BI-GAT & 0.990±0.006 & 0.981±0.015 & 0.671±0.047 & 0.926±0.038 & 0.989±0.009 & 0.911±0.014 \\
GPS-P & 1.000±0.000 & 1.000±0.000 & 0.938±0.038 & 1.000±0.000 & 0.890±0.234 & 0.965±0.049 \\
DI-GPS-P & 1.000±0.000 & 1.000±0.000 & 0.950±0.032 & 0.998±0.006 & 0.994±0.018 & 0.988±0.009 \\
BI-GPS-P & 1.000±0.000 & 1.000±0.000 & 0.960±0.043 & 1.000±0.000 & 0.876±0.196 & 0.967±0.047 \\ \midrule
DGCN & 0.343±0.020 & 0.409±0.028 & 0.122±0.022 & 0.510±0.077 & 0.441±0.088 & 0.365±0.017 \\
DiiGCN & 0.343±0.031 & 0.415±0.036 & 0.116±0.017 & 0.456±0.080 & 0.383±0.092 & 0.342±0.028 \\
MagNet & 0.999±0.003 & 0.999±0.003 & 0.922±0.028 & 0.990±0.014 & 0.966±0.054 & 0.975±0.016 \\ \bottomrule
\end{tabular}}
\vspace{-0.2cm}
\captionof{table}{ID performance on the CG dataset on device `Adreno 640 GPU' with `acc5' metric.}
\label{tab:cg_gpu640_id_acc5}
\end{center}

\vspace{-0.5cm}
\begin{center}
\resizebox{0.8\textwidth}{!}{\begin{tabular}{@{}ccccccc@{}}
\toprule
Method & densenets & mnasnets & mobilenetv3s & mobilenetv4s & nasbench202s & average \\ \midrule
GCN & 0.694±0.011 & 0.747±0.018 & 0.249±0.013 & 0.768±0.061 & 0.957±0.090 & 0.683±0.024 \\
DI-GCN & 0.684±0.027 & 0.717±0.023 & 0.257±0.037 & 0.728±0.068 & 0.941±0.128 & 0.665±0.032 \\
BI-GCN & 0.704±0.022 & 0.757±0.047 & 0.346±0.031 & 0.715±0.107 & 0.712±0.178 & 0.646±0.054 \\
GIN & 1.000±0.000 & 1.000±0.000 & 0.959±0.048 & 0.997±0.006 & 0.635±0.396 & 0.918±0.076 \\
DI-GIN & 1.000±0.000 & 1.000±0.000 & 0.999±0.003 & 1.000±0.000 & 1.000±0.000 & 0.999±0.000 \\
BI-GIN & 1.000±0.000 & 1.000±0.000 & 1.000±0.000 & 1.000±0.000 & 1.000±0.000 & 1.000±0.000 \\
GAT & 0.992±0.010 & 0.991±0.014 & 0.701±0.104 & 0.985±0.009 & 0.985±0.038 & 0.930±0.026 \\
DI-GAT & 0.984±0.015 & 0.956±0.032 & 0.669±0.051 & 0.949±0.043 & 0.675±0.055 & 0.846±0.028 \\
BI-GAT & 1.000±0.000 & 1.000±0.000 & 0.926±0.020 & 0.999±0.003 & 1.000±0.000 & 0.985±0.004 \\
GPS-P & 1.000±0.000 & 1.000±0.000 & 0.998±0.004 & 1.000±0.000 & 0.990±0.022 & 0.997±0.005 \\
DI-GPS-P & 1.000±0.000 & 1.000±0.000 & 0.999±0.003 & 1.000±0.000 & 1.000±0.000 & 0.999±0.000 \\
BI-GPS-P & 1.000±0.000 & 1.000±0.000 & 1.000±0.000 & 1.000±0.000 & 0.986±0.023 & 0.997±0.004 \\ \midrule
DGCN & 0.664±0.024 & 0.722±0.024 & 0.231±0.022 & 0.800±0.061 & 0.699±0.082 & 0.623±0.026 \\
DiiGCN & 0.662±0.031 & 0.703±0.027 & 0.218±0.030 & 0.745±0.067 & 0.617±0.114 & 0.589±0.031 \\
MagNet & 1.000±0.000 & 1.000±0.000 & 0.994±0.007 & 1.000±0.000 & 0.999±0.003 & 0.998±0.001 \\ \bottomrule
\end{tabular}}
\vspace{-0.2cm}
\captionof{table}{ID performance on the CG dataset on device `Adreno 640 GPU' with `acc10' metric.}
\label{tab:cg_gpu_id_acc10}
\end{center}
\vspace{-0.5cm}
\begin{center}
\resizebox{0.8\textwidth}{!}{\begin{tabular}{@{}ccccccc@{}}
\toprule
\multirow{2}{*}{Method} & \multicolumn{3}{c}{proxylessnass} & \multicolumn{3}{c}{resnets} \\ \cmidrule(l){2-7} 
 & rmse & acc5 & acc10 & rmse & acc5 & acc10 \\ \midrule
GCN & 5.600±0.162 & 0.169±0.018 & 0.303±0.018 & 82.509±1.149 & 0.006±0.004 & 0.011±0.004 \\
DI-GCN & 6.163±0.463 & 0.145±0.013 & 0.283±0.022 & 82.785±1.409 & 0.007±0.004 & 0.015±0.004 \\
BI-GCN & 5.039±0.203 & 0.147±0.012 & 0.279±0.021 & 70.904±1.024 & 0.001±0.002 & 0.003±0.004 \\
GIN & 3.704±0.383 & 0.267±0.057 & 0.488±0.069 & 65.961±7.773 & 0.029±0.014 & 0.067±0.027 \\
DI-GIN & 3.672±0.328 & 0.249±0.028 & 0.461±0.044 & 62.820±4.603 & 0.038±0.008 & 0.078±0.017 \\
BI-GIN & 3.326±0.218 & 0.285±0.031 & 0.535±0.042 & 69.651±5.469 & 0.037±0.010 & 0.067±0.016 \\
GAT & 3.646±0.751 & 0.212±0.058 & 0.408±0.089 & 66.092±7.723 & 0.018±0.012 & 0.039±0.029 \\
DI-GAT & 4.893±0.580 & 0.155±0.031 & 0.312±0.048 & 69.358±4.897 & 0.022±0.012 & 0.047±0.022 \\
BI-GAT & 3.666±0.544 & 0.246±0.057 & 0.453±0.080 & 76.832±4.857 & 0.000±0.001 & 0.001±0.003 \\
GPS-P & 2.789±0.093 & 0.293±0.086 & 0.627±0.021 & 74.002±1.594 & 0.006±0.005 & 0.011±0.006 \\
DI-GPS-P & 2.883±0.134 & 0.303±0.053 & 0.605±0.038 & 76.163±2.222 & 0.006±0.004 & 0.016±0.011 \\
BI-GPS-P & 2.747±0.101 & 0.283±0.063 & 0.653±0.020 & 72.210±0.404 & 0.011±0.002 & 0.020±0.005 \\
DGCN & 4.985±0.157 & 0.160±0.016 & 0.300±0.018 & 85.713±1.394 & 0.007±0.007 & 0.011±0.009 \\
DiGCN & 5.362±0.645 & 0.148±0.012 & 0.299±0.020 & 86.605±0.664 & 0.003±0.003 & 0.004±0.005 \\
MagNet & 3.309±0.187 & 0.431±0.030 & 0.594±0.043 & 77.379±1.987 & 0.004±0.006 & 0.005±0.008 \\ \bottomrule
\end{tabular}}
\vspace{-0.2cm}
\captionof{table}{OOD performance on the CG dataset on device `Adreno 640 GPU'.}
\label{tab:cg_gpu640_ood}
\end{center}

\clearpage
\subsection{Comparison between NPE and EPE}
\label{app:result_npe_epe}
\vspace*{-0.5cm}
\begin{center}
\resizebox{\textwidth}{!}{\begin{tabular}{@{}ccccccc@{}}
\toprule
\multirow{2}{*}{Method} & \multicolumn{2}{c}{gain} & \multicolumn{2}{c}{PM} & \multicolumn{2}{c}{BW} \\ \cmidrule(l){2-7} 
 & mse & rmse & mse & rmse & mse & rmse \\ \midrule
BI-GIN+NPE & \textbf{0.135±0.009} & \textbf{0.367±0.012} & 1.296±0.024 & 1.138±0.010 & 19.215±1.044 & 4.382±0.117 \\
BI-GINE+EPE & 0.149±0.009 & 0.386±0.012 & \textbf{1.283±0.033} & \textbf{1.132±0.014} & \textbf{17.399±0.644} & \textbf{4.170±0.076} \\ \midrule
BI-GPS+NPE & 0.122±0.007 & 0.349±0.010 & 1.212±0.058 & 1.100±0.026 & 20.475±8.853 & 4.456±0.825 \\
BI-GPS+EPE & \textbf{0.115±0.008} & \textbf{0.339±0.011} & \textbf{1.206±0.090} & \textbf{1.097±0.040} & \textbf{18.153±2.235} & \textbf{4.253±0.256} \\ \bottomrule
\end{tabular}}
\vspace{-0.2cm}
\captionof{table}{NPE v.s. EPE on ID data on the AMP dataset.}
\label{tab:amp_id_ne}
\end{center}
\vspace{-0.6cm}
\begin{center}

\resizebox{\textwidth}{!}{\begin{tabular}{@{}ccccccc@{}}
\toprule
 & \multicolumn{2}{c}{dsp} & \multicolumn{2}{c}{lut} & \multicolumn{2}{c}{cp} \\ \midrule
method & mse & r2 & mse & r2 & mse & r2 \\ \midrule
BI-GINE+NPE & 2.508±0.183 & 0.977±0.001 & 1.983±0.078 & 0.879±0.006 & 0.617±0.026 & 0.858±0.004 \\
BI-GINE+EPE & \textbf{2.127±0.085} & \textbf{0.981±0.000} & \textbf{1.729±0.096} & \textbf{0.895±0.007} & \textbf{0.607±0.022} & \textbf{0.857±0.007} \\ \midrule
BI-GPS+NPE & 2.442±0.303 & 0.979±0.002 & 2.112±0.216 & 0.873±0.014 & 0.621±0.018 & 0.859±0.009 \\
BI-GPS+EPE & \textbf{2.133±0.148} & \textbf{0.981±0.001} & \textbf{1.957±0.125} & \textbf{0.883±0.011} & \textbf{0.602±0.017} & \textbf{0.861±0.007} \\ \bottomrule
\end{tabular}}
\vspace{-0.2cm}
\captionof{table}{NPE v.s. EPE on ID data on the HLS dataset.}
\label{tab:hls_id_ne}
\end{center}
\vspace{-0.8cm}
\begin{center}
\resizebox{\textwidth}{!}{\begin{tabular}{ccccccccc}
\hline
 & \multicolumn{4}{c}{shared} & \multicolumn{4}{c}{root} \\ \hline
 & accuracy & precision & recall & f1 & accuracy & precision & recall & f1 \\ \hline
BI-GIN+NPE & \textbf{0.999±0.000} & \textbf{0.987±0.039} & \textbf{0.999±0.000} & \textbf{0.991±0.026} & \textbf{0.999±0.000} & \textbf{0.999±0.000} & \textbf{0.999±0.000} & \textbf{0.999±0.000} \\
BI-GIN+EPE & \textbf{0.999±0.001} & 0.974±0.079 & 0.974±0.078 & 0.974±0.079 & \textbf{0.999±0.000} & \textbf{0.999±0.000} & \textbf{0.999±0.000} & \textbf{0.999±0.000} \\ \hline
\end{tabular}}
\vspace{-0.2cm}
\captionof{table}{NPE v.s. EPE on ID data on the SR dataset.}
\label{tab:sr_id_ne}
\end{center}
\vspace{-0.8cm}
\begin{center}
\resizebox{\textwidth}{!}{\begin{tabular}{@{}cccccccccc@{}}
\toprule
 & method & blabla & usb\_cdc\_core & wbqspiflash & cic\_decimator & picorv32a & zipdiv & usb & average \\ \midrule
\multirow{2}{*}{hold rmse} & BI-GINE+NPE & 0.449±0.106 & 0.327±0.161 & 0.162±0.051 & 0.063±0.025 & 2.006±0.115 & 0.107±0.080 & 0.049±0.018 & 0.452±0.040 \\
 & BI-GINE+EPE & \textbf{0.181±0.059} & \textbf{0.021±0.008} & \textbf{0.031±0.006} & \textbf{0.010±0.003} & \textbf{0.652±0.070} & \textbf{0.026±0.011} & \textbf{0.014±0.002} & \textbf{0.134±0.012} \\ \midrule
\multirow{2}{*}{hold R2} & BI-GINE+NPE & 0.995±0.001 & 0.834±0.081 & 0.940±0.018 & 0.928±0.028 & 0.855±0.008 & 0.970±0.022 & 0.928±0.026 & 0.921±0.018 \\
 & BI-GINE+EPE & \textbf{0.998±0.000} & \textbf{0.988±0.004} & \textbf{0.981±0.003} & \textbf{0.993±0.002} & \textbf{0.954±0.004} & \textbf{0.993±0.002} & \textbf{0.986±0.002} & \textbf{0.985±0.001} \\ \midrule
\multirow{2}{*}{setup rmse} & BI-GINE+NPE & 55.889±2.928 & 5.481±1.416 & 13.006±2.716 & 2.438±0.606 & 34.429±2.865 & 12.805±4.583 & 1.728±0.331 & 17.968±1.206 \\
 & BI-GINE+EPE & \textbf{15.134±2.195} & \textbf{1.327±0.747} & \textbf{2.718±1.297} & \textbf{0.966±0.877} & \textbf{17.996±3.227} & \textbf{6.436±5.416} & \textbf{0.885±0.317} & \textbf{6.494±1.449} \\ \midrule
\multirow{2}{*}{setup R2} & BI-GINE+NPE & 0.142±0.044 & -0.021±0.264 & 0.326±0.140 & -0.149±0.285 & 0.173±0.068 & 0.364±0.227 & 0.306±0.132 & 0.163±0.093 \\
 & BI-GINE+EPE & \textbf{0.767±0.033} & \textbf{0.752±0.139} & \textbf{0.859±0.067} & \textbf{0.544±0.413} & \textbf{0.567±0.077} & \textbf{0.680±0.268} & \textbf{0.644±0.127} & \textbf{0.688±0.091} \\ \bottomrule
\end{tabular}}
\vspace{-0.2cm}
\captionof{table}{NPE v.s. EPE on ID data on the TIME dataset.}
\label{tab:time_id_ne}
\end{center}
\vspace*{-0.5cm}
\begin{center}
\resizebox{\textwidth}{!}{\begin{tabular}{@{}cccccccc@{}}
\toprule
 & Method & densenets & mnasnets & mobilenetv2s & mobilenetv3s & nasbench201s & average \\ \midrule
\multirow{2}{*}{cpu rmse} & BI-GIN+NPE & \textbf{7.734±0.602} & 2.053±0.235 & 3.788±0.424 & 1.590±0.371 & 0.725±0.173 & 3.178±0.168 \\
 & BIGINE+EPE & 7.550±0.291 & \textbf{1.728±0.224} & \textbf{3.064±0.326} & \textbf{1.176±0.155} & \textbf{0.419±0.033} & \textbf{2.788±0.147} \\ \midrule
\multirow{2}{*}{cpu acc5} & BI-GIN+NPE & 0.893±0.014 & 0.821±0.056 & 0.653±0.052 & 0.705±0.153 & 0.545±0.351 & 0.723±0.104 \\
 & BIGINE+EPE & \textbf{0.907±0.017} & \textbf{0.873±0.042} & \textbf{0.766±0.041} & \textbf{0.861±0.036} & \textbf{0.901±0.029} & \textbf{0.861±0.019} \\ \midrule
\multirow{2}{*}{cpu acc10} & BI-GIN+NPE & 0.999±0.003 & 0.994±0.006 & 0.940±0.025 & 0.907±0.066 & 0.940±0.060 & 0.956±0.028 \\
 & BIGINE+EPE & \textbf{1.000±0.000} & \textbf{0.999±0.003} & \textbf{0.969±0.023} & \textbf{0.994±0.009} & \textbf{0.999±0.003} & \textbf{0.992±0.006} \\ \midrule
\multirow{2}{*}{gpu630 rmse} & BI-GIN+NPE & 0.440±0.020 & 0.112±0.009 & \textbf{0.254±0.019} & 0.099±0.014 & \textbf{0.056±0.006} & 0.192±0.006 \\
\bottomrule
\end{tabular}}
\vspace{-0.2cm}
\captionof{table}{NPE v.s. EPE on ID data on the CG dataset.}
\label{tab:cg_id_ne}
\end{center}
\vspace*{-1.6cm}

\clearpage

\clearpage

\vspace{-1cm}
\begin{center}
\resizebox{\textwidth}{!}{\begin{tabular}{ccccccc}
\hline
\multirow{2}{*}{Method} & \multicolumn{2}{c}{gain} & \multicolumn{2}{c}{PM} & \multicolumn{2}{c}{BW} \\ \cline{2-7} 
 & mse & rmse & mse & rmse & mse & rmse \\ \hline
BI-GIN+NPE & 0.303±0.046 & 0.549±0.042 & 1.379±0.027 & 1.174±0.011 & 25.967±1.646 & 5.093±0.161 \\
BI-GINE+EPE & \textbf{0.302±0.037} & \textbf{0.549±0.033} & \textbf{1.373±0.005} & \textbf{1.171±0.002} & \textbf{22.339±1.413} & \textbf{4.724±0.150} \\ \hline
BI-GPS+NPE & 0.314±0.030 & 0.560±0.027 & 1.315±0.050 & 1.146±0.021 & 26.607±10.277 & 5.087±0.897 \\
BI-GPS+EPE & \multicolumn{1}{l}{\textbf{0.302±0.071}} & \multicolumn{1}{l}{\textbf{0.546±0.060}} & \textbf{1.314±0.143} & \textbf{1.126±0.060} & \textbf{21.815±1.973} & \textbf{4.666±0.206} \\ \hline
\end{tabular}}
\vspace{-0.2cm}
\captionof{table}{NPE v.s. EPE on OOD data on the AMP dataset.}
\label{tab:amp_ood_ne}
\end{center}
\begin{center}
\resizebox{\textwidth}{!}{\begin{tabular}{@{}ccccccc@{}}
\toprule
 & \multicolumn{2}{c}{dsp} & \multicolumn{2}{c}{lut} & \multicolumn{2}{c}{cp} \\ \midrule
method & mse & r2 & mse & r2 & mse & r2 \\ \midrule
BI-GINE+NPE & 3.434±0.238 & 0.964±0.002 & 0.113±0.019 & 0.971±0.004 & \textbf{0.450±0.013} & \textbf{0.830±0.008} \\
BI-GINE+EPE & \textbf{3.243±0.098} & \textbf{0.966±0.000} & \textbf{0.102±0.019} & \textbf{0.973±0.005} & \textbf{0.452±0.022} & \textbf{0.823±0.011} \\ \midrule
BI-GPS+NPE & 3.209±0.263 & 0.967±0.001 & 0.133±0.027 & 0.968±0.006 & 0.496±0.017 & 0.812±0.016 \\
BI-GPS+EPE & \textbf{3.205±0.026} & \textbf{0.968±0.001} & \textbf{0.102±0.017} & \textbf{0.972±0.003} & \textbf{0.474±0.017} & \textbf{0.830±0.006} \\ \bottomrule
\end{tabular}}
\vspace{-0.2cm}
\captionof{table}{NPE v.s. EPE on OOD data on the HLS dataset.}
\label{tab:hls_ood_ne}
\end{center}
\begin{center}
\resizebox{\textwidth}{!}{\begin{tabular}{@{}ccccccccc@{}}
\toprule
 & \multicolumn{4}{c}{shared} & \multicolumn{4}{c}{root} \\ \midrule
 & accuracy & precision & recall & f1 & accuracy & precision & recall & f1 \\ \midrule
BI-GIN+NPE & 0.712±0.027 & \textbf{0.510±0.103} & 0.591±0.021 & 0.502±0.032 & 0.696±0.057 & 0.556±0.066 & \textbf{0.616±0.072} & \textbf{0.567±0.068} \\
BI-GINE+EPE & \textbf{0.725±0.037} & \textbf{0.510±0.059} & \textbf{0.604±0.030} & \textbf{0.530±0.038} & \textbf{0.747±0.035} & \textbf{0.569±0.113} & 0.508±0.080 & 0.520±0.090 \\ \bottomrule
\end{tabular}}
\captionof{table}{NPE v.s. EPE on OOD data on the SR dataset.}
\label{tab:sr_ood_ne}
\end{center}
\begin{center}
\resizebox{\textwidth}{!}{\begin{tabular}{@{}cccccc@{}}
\toprule
 & \multirow{2}{*}{Method} & \multicolumn{2}{c}{xtea} & \multicolumn{2}{c}{synth\_ram} \\ \cmidrule(l){3-6} 
 &  & mse & r2 & mse & r2 \\ \midrule
\multirow{2}{*}{hold} & BI-GINE+NPE & 6.936±0.914 & -0.092±0.144 & 0.743±0.313 & 0.798±0.084 \\
 & BI-GINE+-EPE & \textbf{2.074±0.474} & \textbf{0.673±0.074} & \textbf{0.617±0.906} & \textbf{0.832±0.245} \\ \midrule
\multirow{2}{*}{setup} & BI-GINE+NPE & 98.690±55.964 & -2.326±1.886 & 629.630±107.437 & -1215.849±207.638 \\
 & BI-GINE+-EPE & \textbf{59.401±13.573} & \textbf{-1.002±0.457} & \textbf{619.030±136.176} & \textbf{-1195.363±263.179} \\ \bottomrule
\end{tabular}}
\captionof{table}{NPE v.s. EPE on OOD data on the TIME dataset.}
\label{tab:time_ood_ne}
\end{center}
\begin{center}
\resizebox{\textwidth}{!}{\begin{tabular}{@{}cccccccc@{}}
\toprule
 & \multirow{2}{*}{Method} & \multicolumn{3}{c}{proxylessnass} & \multicolumn{3}{c}{resnets} \\ \cmidrule(l){3-8} 
 &  & rmse & acc5 & acc10 & rmse & acc5 & acc10 \\ \midrule
cpu & BI-GIN+NPE & 15.045±5.642 & 0.264±0.138 & 0.487±0.199 & 388.079±36.250 & \textbf{0.041±0.014} & \textbf{0.073±0.024} \\
 & BI-GINE+EPE & \textbf{11.049±0.909} & \textbf{0.310±0.024} & \textbf{0.588±0.049} & \textbf{381.432±28.102} & 0.036±0.009 & 0.072±0.016 \\ \midrule
gpu630 & BI-GIN+NPE & \textbf{2.843±0.289} & 0.277±0.031 & 0.538±0.055 & \textbf{97.056±7.138} & \textbf{0.034±0.012} & \textbf{0.065±0.020} \\
 & BI-GINE+EPE & 2.947±0.239 & \textbf{0.370±0.082} & \textbf{0.595±0.051} & 108.959±2.594 & 0.020±0.011 & 0.033±0.018 \\ \cmidrule(l){2-8} 
gpu640 & BI-GIN+NPE & 3.591±0.286 & \textbf{0.303±0.062} & \textbf{0.549±0.050} & 76.590±4.263 & 0.024±0.009 & 0.048±0.011 \\
 & BI-GINE+EPE & \textbf{3.419±0.266} & 0.295±0.034 & 0.525±0.050 & \textbf{68.530±3.603} & \textbf{0.037±0.020} & \textbf{0.065±0.023} \\ \bottomrule
\end{tabular}}
\captionof{table}{NPE v.s. EPE on OOD data on the CG dataset.}
\label{tab:cg_ood_ne}
\end{center}

\clearpage


\end{document}